\documentclass[lettersize,journal]{IEEEtran}
\usepackage{amsmath,amsfonts}
\usepackage{algorithmic}
\usepackage{algorithm}
\usepackage{array}
\usepackage{textcomp}
\usepackage{stfloats}
\usepackage{url}
\usepackage{verbatim}
\usepackage{graphicx}
\usepackage{subfigure}
\usepackage{cite}
\newtheorem{theorem}{Theorem}

\newtheorem{assumption}{Assumption}
\newtheorem{definition}{Definition}
\newtheorem{remark}{Remark}
\allowdisplaybreaks[4]
\begin{document}

\title{How global observation works in Federated Learning: Integrating vertical training into Horizontal Federated Learning}

\author{Shuo Wan,
	    Jiaxun Lu,~\IEEEmembership{Member,~IEEE,}
	    Pingyi Fan,~\IEEEmembership{Senior Member,~IEEE,}
	    Yunfeng Shao,~\IEEEmembership{Member,~IEEE,}
	    Chenghui Peng, and
	    Khaled B. Letaief,~\IEEEmembership{Fellow,~IEEE}
        % <-this % stops a spaceh7y
}

% The paper headers

% Remember, if you use this you must call \IEEEpubidadjcol in the second
% column for its text to clear the IEEEpubid mark.

\maketitle

\begin{abstract}
Federated learning (FL) has recently emerged as a transformative paradigm that jointly train a model with distributed data sets in IoT while avoiding the need for central data collection. Due to the limited observation range, such data sets can only reflect local information, which limits the quality of trained models. In practice, the global information and local observations would require a joint consideration for learning to make a reasonable policy. However, in horizontal FL, the central agency only acts as a model aggregator without utilizing its global observation to further improve the model. This could significantly degrade the performance in some missions such as traffic flow prediction in network systems, where the global information may enhance the accuracy. Meanwhile, the global feature may not be directly transmitted to agents for data security. How to utilize the global observation residing in the central agency while protecting its safety thus rises up as an important problem in FL. In this paper, we develop a vertical-horizontal federated learning (VHFL) process, where the global feature is shared with the agents in a procedure similar to that of vertical FL without any extra communication rounds. By considering the delay and packet loss, we will analyze VHFL convergence and validate its performance by experiments. It is shown that the proposed VHFL could enhance the accuracy compared with horizontal FL while still protecting the security of global data.
\end{abstract}


\section{Introduction}
\IEEEPARstart{N}{owadays}, big data is typically collected from distributed facilities or agents. Due to limitation on the network capacity and data security, however, it is hard to process data for central analysis. Therefore, distributed learning rises up as an important issue for data analysis in large-scale networks. In \cite{mcmahan2017communication}, the authors proposed Horizontal federated learning (HFL) which jointly trains a model among distributed clients through local training and model aggregation. In this process, the clients' models are periodically trained based on previously aggregated models until convergence. Such a learning scheme allows applications to collectively reap the benefits of shared models trained from the rich data while avoiding the need of central data collection. That is, the trained FL model could assist the policy decision in IoT applications with a wider learned experience while protecting the security of the original agents.

Although the combined experience from distributed agents will largely improve the performance, it is still not enough if the global observation is not taken advantage of. Also and since the agents in IoT rarely work independently, correlations among them is unavoidable, which stresses the importance of global information in policy decision. For instance, the network trafic flow in one community will also affect its neighborhood. The self-driving performance may count on the positions and directions of neighboring cars. Such information typically resides in a central agency. Hence, if we can properly use such global observation in federated training, then the learned model will definitely give a better policy. Therefore, to improve FL for intelligent decisions among clients, how to utilize the global observation in the training procedure is a critical issue.

Integrating the global feature into horizontal FL faces many practical challenges in network system. Since the security may be threatened if the global observations are broadcast, the central agency often needs to keep its global data from being directly accessed by distributed clients. Therefore, the processing of global features should reside at the center. In this context and compared with conventional HFL, the central agency should announce its processed results other than just aggregating the model weights. Multiple challenging questions will thus need to be addressed. These include: How should we design such an FL mode to integrate a global processing? How can we jointly train the federated model and the global model with communication-efficient interactions? Does the joint training converges and what are the factors which can influence the convergence? In this paper, we will try to answer these important questions. 

Vertical federated learning (VFL) \cite{hardy2017private,yang2019federatedm} is a modified scheme for cases in which the data sets share the same sample ID space but differ in the feature space. For instance, a commercial company and a bank may seperately record the related features of many users in a city. Combining these features in training will largely enhance the model accuracy. To deal with the security problem, VFL is designed to securely aggregate the varied features and compute the training loss and gradients to build a model with data from both companies collaboratively. 

The training procedure of VFL provides a way to aggregate features from different agents. To integrate the global feature into HFL, the central agency could view the federal clients as another companion. In this sense, combining VFL into horizontal federated training provides a natural way for the central agency to share the global feature in security. It is known that the center in HFL only acts as a model aggregator. Under the concept of VFL, we will add a global model at the center to process the globally observed features. The processed results will be broadcast to distributed clients together with the aggregated model weights. When the local training is completed, the resulting loss for the global model will also be uploaded together with the updated weights of the local models. In this procedure, the extra transmissions are included in the periodic model aggregation and broadcast as in HFL. Since the processing results and feedback loss are rather small compared with the model weights, the system could naturally ensure communication efficiency. In this paper, we will propose a Vertical-Horizontal Federated Learning (VHFL) scheme and design the collaborative training procedure therein. Through theoretical analysis, we will prove its convergence considering the background packet loss and transmission delay in the networking system. Based on the network data among distributed communities, we will also test the convergence of VHFL and verify its performance. Hence, for IoT applications counting on global observations, it will be a practical scheme for distributed learning.

\subsection{Related works}
Subsequent to the work in \cite{mcmahan2017communication}, FL has been widely investigated with improvement and further developments \cite{hardy2017private,yang2019federatedm,li2018federated,sahu2018convergence,reddi2020adaptive,sattler2020clustered,karimireddy2020scaffold}. For vertically partitioned data in distributed clients, \cite{hardy2017private} introduced the VFL scheme with privacy preserving. In \cite{yang2019federatedm}, the authors discussed the concept of federated transfer learning (FTL) for scenarios where two datasets differ both in the samples and feature spaces. This mode is specifically fit for scenarios where the united data sets among clients are very limited for FL. The training mode of FL is challenged by the heterogeneity among clients. Thus, in \cite{li2018federated} and \cite{sahu2018convergence}, the authors proposed the FedProx framework that allows each participating client to perform a variety amount of local training to achieve a more robust convergence. With the same aim, \cite{reddi2020adaptive} tried to apply an adaptive optimizer on the model updates in FL. To cope with the non-i.i.d. data sets among clients, \cite{sattler2020clustered} presented clustered FL (CFL) which exploits the geometric properties of the FL loss surface to group the client population into clusters with jointly trainable data distributions. By counting on the client set for solution, \cite{karimireddy2020scaffold} tried to maintain a control variate to correct the client-drift in its local updates. Among these works, \cite{hardy2017private,yang2019federatedm} tried to propose a new training mode for specific application scenarios while \cite{li2018federated,sahu2018convergence,reddi2020adaptive,sattler2020clustered,karimireddy2020scaffold} tried to improve the original FedAvg for better convergence.

Since deploying distributed learning over wireless
networks faces several challenges including the uncertain wireless environment, limited
wireless resources and hardware resources \cite{chendistributed}, researchers discussed the facility basis of the platform \cite{wang2019adaptive,LiDelay,sattler2019robust,yangbnn,chen2020joint,wan2021convergence,liu2020client,LimHierarchical}. In \cite{wang2019adaptive}, the authors considered the capability of computing and model transmission and proposed a strategy to schedule the model uploading slots. In \cite{LiDelay}, the authors studied the delay in FL tasks and presented a unified framework to estimate the delay through fading channels. To relieve the transmission burden in FL, \cite{sattler2019robust} proposed the sparse ternary compression (STC) framework to model transmission in FL and \cite{yangbnn} considered training a binary neural network in FL. Considering the packet loss in wireless transmission, \cite{chen2020joint} analyzed the FL convergence with improved strategies for system scheduling. By analyzing the effects of hyper-parameters on FL convergence, \cite{wan2021convergence} proposed theoretical design principles for FL implementation with the corresponding scheduling strategies. Considering large-scale FL, the authors of \cite{liu2020client} proposed hierarchical FL with edge servers for intermediate aggregation and \cite{LimHierarchical} discussed the edge association and resource allocation in this system.

\subsection{Contributions}
In this paper, we focus on how to utilize the global observations in the central agency to assist FL on distributed data sets so that the learned model could achieve better performance. Our contributions can be summarized as follows.
\begin{enumerate}
	\item We integrate the global observations into HFL and propose the Vertical-Horizontal Federated Learning (VHFL) scheme. We develop a collaborative training procedure with almost equivalent costs in communication compared with conventional HFL. In practice, it will be important for applications counting on the global information with security concern.
	\item Through theoretical analysis, we prove the convergence of VHFL with analytic bounds and clearly demonstrates the effects of system parameters on training. By network analysis, we analyze the theoretical link between the network delay and transmission failure. Combined with the convergence bound, we also propose principles to set the delay due to network conditions and training requirements of VHFL. These results are expected to provide theoretical insights for the implementation of VHFL in future 6G networks.
	\item We test VHFL on the measured data from a real network among communities and observe an improved accuracy in the prediction of network flow compared with HFL. It will be shown that for such applications among collaborative clients, the global information can significantly enhance the performance as clearly reflected by experiments.
\end{enumerate}
\begin{figure}[tbp]
	\centering
	\includegraphics[width=0.99\columnwidth]{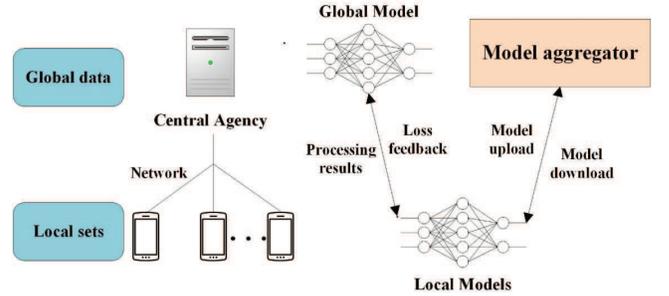}
	\caption{The Vertical-Horizontal Federated Learning (VHFL) system structure.}\label{VHFL_sys}
\end{figure}
\section{Vertical-Horizontal Federated Learning}
We denote the feature space as $X$, the label space as $Y$ and the client sample ID space as $I$. The client set is $\mathbb{N}=\{1,2,\dots,N\}$. Therein, client $j (j\in N)$ stores the data set $\mathbb{D}_{j}=(X_{j}, Y_{j}, I_{j})$, where $X_{j}$ and $Y_{j}$ are the feature space and label space of client $j$ and $I_{j}$ with $|I_{j}|=D_{j}$ is the index set of samples in client $j$. Denoting the model weight as $w$, then the training loss of client $j$ is
\begin{equation}\label{local_loss}
	f_{j}(w)=\frac{1}{D_{j}}\sum_{m=1}^{D_{j}}l(w,(x_{i_{m}},y_{i_{m}})),
\end{equation}
where $i_{m} \in I_{j}$ is the sample index of client $j$ and $l(\cdot ,\cdot )$ is the loss function. In this section, we will first introduce the system model and background learning techniques. Then we will specifically discuss the training procedure of VHFL in detail.

\subsection{Problem formulation and System model}\label{basic_model}
In this paper, we consider a collaborative learning system as shown in Fig. \ref{VHFL_sys}. In the network system, the intelligent clients $\mathbb{N}=\{1, 2, \dots, N\}$ reside in distributed regions, each with its own data set $\mathbb{D}_{j}=(X_{j}, Y_{j}, I_{j})$ from local observations. The central agency should be a network center which stores the synthesized information. The global data set at the center is denoted as $\mathbb{D}_{0}=(X_{0}, I_{0})$. Sometimes, if the global observations are obtained generally not corresponding to the specific sampling ID, the global data will only involve the observing features in the form $\mathbb{D}_{0}=(X_{0})$, which can be viewed as a special case.

Assuming that the clients intend to train a model for policy making in local applications, then they will need the combined experience from other clients and the central agency. Samples of other clients could enrich the model training with improved accuracy. Meanwhile, the global feature contains observations of the whole system, which is often crucial for policy decisions. Here comes the major problem. Hence how the clients and central agency could train a model collaboratively without transmitting their data sets is a critical question. Other questions include: Could the training ensure a convergence in the network system? Considering the network connection, how should we implement the joint training considering unreliable communication?

As shown in Fig. \ref{VHFL_sys}, the central agency and clients are connected through the network, which can be wireless or wired. Each client trains the model locally and the center also maintains a global model to process the central set $\mathbb{D}_{0}$. The center and local agents interact periodically for model aggregation and vertical training.

Note that it is not realistic for the network to specifically support the learning system. All services should go through a public network for communication. Considering the random arriving rate and transmission rate naturally existing in the public network, the communication between a client and a central agency should be modeled as a queuing system. In this paper, we will analyze the delay and packet loss of VHFL considering an M/G/1 queuing network. The packet arriving rate is $\lambda_{N}$, which can be measured by experiment. The transmission rate depends on the network conditions. During busy periods, the network is rather congested while the service at casual preriods can be much higher. This is a very natural property of the public network which can be modeled by a 2-stage hyper-exponential distribution. Therein, the probability distribution of the service time can be given by
\begin{equation}\label{delay_prob}
	B(t)=\alpha_{1}\mu_{1}e^{-\mu_{1}t}+\alpha_{2}\mu_{2}e^{-\mu_{2}t},
\end{equation}
where $\mu_{1}$, $\mu_{2}$ are seperately the service rate at the idle and busy states ($\mu_{1}>\mu_{2}$) and $\alpha_{1}$, $\alpha_{2}$ are the probability weights of the two states. These parameters can be directly measured from the practical flow records in the network for guidence.

In HFL or VHFL, the center will collect messages from clients for central training or weight aggregation. Thus, the varied delays will be an important factor to disturb the central collection. In practical operations, the center will set a deadline and any uploading beyond the deadline will be lost. Note that the TCP will retransmit lost packets as a result of congestion. From the view of the central collector, this is equivalent to a larger delay modeled by the congested rate $\mu_{2}$. Therefore, the packet loss due to time out could cover most cases in the network. In this paper, we will analyze the uploading delay to guide the system design for VHFL implementation. 

\subsection{Background technique: Federated Learning}\label{FLBG}
\subsubsection{Horizontal FL (HFL)}
The HFL system involves a center and a set of clients $\mathbb{N}$. For an arbitrary client $i$ and $j$ $(i \neq j)$, the stored data sets satify $X_{i} = X_{j}$, $Y_{i}=Y_{j}$ and $I_{i} \neq I_{j}$. That is, the clients in HFL share the same space for their features and labels in a common training task. However, due to different positions, the sampling ID naturally varies among the clients. Thus, HFL must share the sample sets with varied ID spaces to train a general model by local training and model aggregation. Given the local loss function, $f_{j}(w)$, as defined in (\ref{local_loss}) and $q_{j}$ as the weight of client $j$, the training loss of HFL is defined as
\begin{equation}\label{HFL_loss}
	f(w)=\sum_{j=1}^{N}q_{j}f_{j}(w).
\end{equation}
The training process is done periodically with $E_{l}$ as the period length. Defining $t$ as an arbitrary training slot, $t_{c}=\left \lfloor \frac{t}{E_{l}} \right \rfloor E_{l}$ is the start slot of the current global epoch. At $t_{c}$, clients download the updated model $\bar{w}^{t_{c}}$ to initialize the upcoming local training. Then at $t \neq t_{c}$, each client is trained locally as
\begin{equation}\label{local_update}
	w_{j}^{t_{c}+i+1} = w_{j}^{t_{c}+i}-\eta^{t_{c}+i}\bigtriangledown f_{j}(w_{j}^{t_{c}+i}),
\end{equation}
where $w_{j}^{t}$ is the local model weight of client $j$ at slot $t$ and $\eta^{t}$ is the learning rate. The stochastic gradient decent in (\ref{local_update}) can be varied with other optimizers, following the same procedure. After $E_{l}$ epochs of local training, a client set $\mathbb{P}$ ($|\mathbb{P}|=K$, $K \leq N$) out of $\mathbb{N}$ will be selected to upload the locally trained model for aggregation at the center. Upon receiving the models, the central agency will update the federal model weight as
\begin{equation}\label{aggregate_model}
	\bar{w}^{t} = \frac{N}{K}\sum_{j \in \mathbb{P}} q_{j}w_{j}^{t},
\end{equation}
where $E[\frac{N}{K}\sum_{j\in \mathbb{P}}q_{j}]=1$. Then $\bar{w}^{t}$ will be downloaded by clients to initialize the local training in the next global epoch. Such HFL learning procedure requires clients to share the same space in feature and label so that they could train on the same model structure. By aggregation, the models trained by distributed sampling sets could be combined as the federal model with the experience of all the clients. Its convergence has been proved by experiments and theoretical analysis \cite{mcmahan2017communication,wan2021convergence}. 

\subsubsection{Vertical FL (VFL)}
In some cases, two agents may share the same sample space ($I_{A}=I_{B}$) but differ in the feature spaces ($X_{A} \neq X_{B}$) \cite{yang2019federatedm}. For instance, a bank $A$ and an e-commerce company $B$ may share the same user set containing most residents of a particular location. However, due to the difference in their business, the two companies naturally record different features of a user. Suppose that two agents want to have a prediction model of product purchases based on the recorded features in both data bases, then VFL will satisfy the requirement while avoiding the need of data collection. 

In VFL, we have $X_{A}\neq X_{B}$, $Y_{A} \neq Y_{B}$ and $I_{A}=I_{B}$. Let us suppose company $B$ contains labels for the prediction task. Thus, the data sets of both companies are $\{x_{i}^{A}\}$ and $\{x_{i}^{B}, y_{i}\}$. Denoting the model weight as $w_{A}$ and $w_{B}$, and the learning rate as $\eta$, then the training objective is
\begin{equation}\label{training_obj}
	\underset{w_{A}, w_{B}}{{\rm min}} f(w_{A}, w_{B})=\underset{w_{A}, w_{B}}{{\rm min}} \sum_{i} L((w_{A}, x_{i}^{A}), (w_{B}, x_{i}^{B}), y_{i}),
\end{equation}
where $L((w_{A}, x_{i}^{A}), (w_{B}, x_{i}^{B}), y_{i})$ is the vertical model loss with joint feature processing of sample $i$. Since the label is stored in company $B$, the model in $A$ will process $x_{i}^{A}$ and output $u_{i}^{A}=h_{w_{A}}(x_{i}^{A})$ to model $B$. Then model $B$ will take $u_{i}^{A}$ and $x_{i}^{B}$ as input to obtain the prediction result as $u_{i}^{B}=h_{w_{B}}(u_{i}^{A}, x_{i}^{B})$. By taking the Mean Square Error (MSE) loss as an example, then the loss function in (\ref{training_obj}) can be given as
\begin{equation}\label{VFL_loss}
	L((w_{A}, x_{i}^{A}), (w_{B}, x_{i}^{B}), y_{i})=||h_{w_{B}}(h_{w_{A}}(x_{i}^{A}), x_{i}^{B})-y_{i}||^{2}.
\end{equation}
For neural networks, model $B$ could process $x_{i}^{B}$ with the former layers and the abstracted result could be combined with $u_{i}^{A}$ to get the prediction result through the final dense layers. For simplicity, we can also process $x_{i}^{B}$ with model $B$ to get $u_{i}^{B}$ and take $u_{i}^{A}+u_{i}^{B}$ as the prediction result \cite{yang2019federatedm}, which is actually a special case of (\ref{VFL_loss}). In the training procedure, $w_{B}$ is updated by
\begin{equation}\label{wB_update}
	w_{B}^{t+1}=w_{B}^{t}-\eta^{t} \bigtriangledown_{w_{B}}f(w_{A}, w_{B}).
\end{equation}
Meanwhile, model $A$ will receive $\bigtriangledown_{u^{A}}f(w_{A}, w_{B})$ for the samples and update $w_{A}$ as
\begin{equation}\label{wA_update}
	w_{A}^{t+1}=w_{A}^{t}-\eta^{t}\bigtriangledown_{w_{A}}\bigtriangledown_{u^{A}}f(w_{A}, w_{B}).
\end{equation}
In terms of neural network implementation, model $A$ will send the output $u^{A}$ to model $B$ and receive the gradient $\bigtriangledown_{u^{A}}f$ to update $w_{A}$ in the backward mode. For security concern, VFL will also apply encryption in transmission and follows the same procedure.

\subsection{Vertical-Horizontal FL (VHFL)}\label{vhfl_tr}
\begin{figure}[tbp]
	\centering
	\includegraphics[width=0.99\columnwidth]{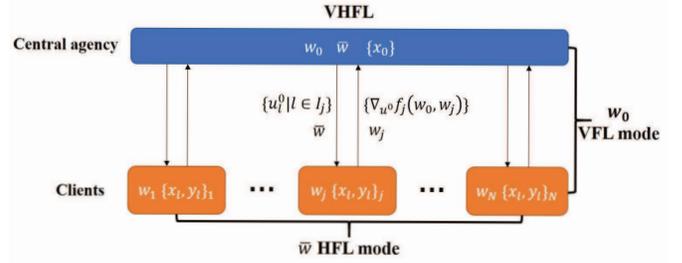}
	\caption{The data flow and interaction mode in the VHFL training procedure.}\label{VHFL_dataflow}
\end{figure}
Due to concern of data privacy and communication burden, an arbitrary client $i$ will not share the local set $\mathbb{D}_{i}$ for central training. Meanwhile, the global observations typically contain sensitive information of the whole system. Spreading the global data around clients could seriously threaten the security. 
Thus, the center will also train a model to process the global information and share the results with clients for prediction. In VHFL, the clients will jointly train in the HFL mode. In this process, the center will not only aggregate the model weights as in HFL, but also train its own global model to assist the HFL among clients with global information. In Section \ref{FLBG}, we have discussed the training procedure of HFL and VFL. Here, we will introduce how these modes are combined for collaborative training in VHFL while avoiding the transmission of raw data.

The data flow and interaction mode in VHFL is shown in Fig. \ref{VHFL_dataflow}. The center maintains a global model with weight $w_{0}$ to process the global features. The distributed clients train the local models $\{w_{j} | j\in{1, 2, \dots, N}\}$ for processing of local features. As in HFL, the center will choose $K$ clients to upload the local weight at each global epoch. The aggregated weights will be transmitted to all clients for processing of their local features. Thus, the federal weight for processing local features is
\begin{equation}\label{wbar_vhfl}
	\bar{w}^{t}=\frac{N}{K}\sum_{j\in \mathbb{P}}q_{j}w_{j}^{t},
\end{equation}
which is similar to (\ref{aggregate_model}) in HFL. Here, the center processes the central data $\mathbb{D}_{0}$ with model $w_{0}$. Assume that the global and local data corresponding to the local set $(X_{j}, Y_{j}, I_{j})$ are $\{x_{l}^{0}\}$ and $\{x_{l}, y_{l}\}$ $(l\in I_{j})$, then the globally processed result transmitted to client $j$ is
\begin{equation}\label{cen_pros}
	u_{l}^{0}=h_{w_{0}}(x_{l}^{0}), l\in I_{j}.
\end{equation}
Upon receiving the updated $\{u_{l}^{0}\}$, the client will apply local model $w_{j}$ to process $\{u_{l}^{0}\}$ and $\{x_{l}\}$ with the final prediction result as
\begin{equation}
	u_{l}=h_{w_{j}}(u_{l}^{0}, x_{l}).
\end{equation}
Taking MSE loss as an example, then the loss function should be 
\begin{equation}
	L((w_{0}, x_{l}^{0}), (w_{j}, x_{l}), y_{l})=||h_{w_{j}}(h_{w_{0}}(x_{l}^{0}), x_{l})-y_{l}||^{2}.
\end{equation}
Thus, the loss function at client $j$ is
\begin{equation}
	f_{j}(w_{0}, w_{j})=\sum_{l\in I_{j}} L((w_{0}, x_{l}^{0}), (w_{j}, x_{l}), y_{l}).
\end{equation}
Referring to (\ref{HFL_loss}) in HFL, the global training objective of VHFL should be
\begin{equation}
	\underset{w_{0}, \bar{w}}{{\rm min}} f(w_{0}, \bar{w})=\underset{w_{0}, \bar{w}}{{\rm min}} \sum_{j=1}^{N}q_{j}f_{j}(w_{0}, \bar{w}),
\end{equation} 
where $q_{j}$ is the weight of client $j$ with $\sum_{j=1}^{N}q_{j}=1$. In VHFL, $w_{0}$ is trained in the VFL mode while $\bar{w}$ is trained in the HFL mode. Similar to HFL, the training procedure of VHFL is performed periodically with $E_{l}$ as the period length as summarized in Algorithm \ref{vhfltp}. Let $t$ denote an arbitrary training slot, $t_{c}=\left \lfloor \frac{t}{E_{l}} \right \rfloor E_{l}$ should be the initial slot of the current global epoch. At $t_{c}$, client $j$ ($j\in \mathbb{N}$) will download the updated $\bar{w}^{t_{c}}$ from the center for local processing. Meanwhile, the central processed features $\{u_{l}^{0}|l\in I_{j}\}$ will also be transmitted to client $j$ simutaneously. Then, client $j$ will train $w_{j}^{t}$ locally as
\begin{equation}\label{VHFL_local_train}
	w_{j}^{t_{c}+i+1}=w_{j}^{t_{c}+i}-\eta^{t_{c}+i}\bigtriangledown_{w_{j}} f_{j}(w_{0}, w_{j}^{t_{c}+i}),
\end{equation}
where $w_{j}^{t_{c}+i}$ is the local model weight of client $j$ at local epoch $i$ and $\eta^{t_{c}+i}$ is the corresponding learning rate. Here the central output $u_{l}^{0}$ from $w_{0}$ is fixed as an input throughout the local epochs in $E_{l}$. Meanwhile, the gradient $\{\bigtriangledown_{u_{l}^{0}}f_{j}(w_{0}, w_{j}^{t})| l\in I_{j}\}$ for each local epoch will be recorded in client $j$. After averaging among the local epochs, the vertical gradient $\{\bigtriangledown_{u_{l}^{0}}f_{j}(w_{0}, w_{j})| l\in I_{j}\}$ will be uploaded to the central agency together with the updated $w_{j}$. Upon receiving the uploaded message, $\{w_{j}|j \in \mathbb{P}\}$ will be updated as (\ref{wbar_vhfl}). At the same time, the center will gather the vertical gradients from all the clients in batches as $\{\bigtriangledown_{u^{0}}f_{j}(w_{0}, w_{j})| j\in \mathbb{P}\}$. Together with the central feature $\{x^{0}\}$, it is natural to see that the stochastic gradient for $w_{0}$ can be directly obtained by backward propagation. Thus, $w_{0}$ is trained centrally as
\begin{equation}\label{vhfl_w0}
	w_{0}^{t_{c}+E_{l}}=w_{0}^{t_{c}}-\eta_{0}^{t_{c}}\sum_{j \in \mathbb{P}}\bigtriangledown_{w_{0}}\bigtriangledown_{u^{0}}f_{j}(w_{0}, w_{j}),
\end{equation}
where $\eta_{0}^{t_{c}}$ is the learning rate for central training. Note that the central operation only occurs at the end of each global epoch, which means (\ref{vhfl_w0}) is performed every $E_{l}$ slots. After training, the renewed $\{u_{l}^{0}\}$ will be downloaded together with the aggregated $\bar{w}$. The local training of the next global epoch then begins. The training procedure is supposed to end when the loss converges. 

The trained model can assist the local applications among clients such as predicting for policy decision. In the inference procedure, the global model will apply the trained $w_{0}$ to process the corresponding global observations. The local model will apply the trained federal weights $\bar{w}$ to process the local observations and central output with the final results. Note that the central observations typically reflect the states of the whole system. They are rather stable and will not change in a period. Therefore, the central agency could process the related central observations for a client in a period without frequent interactions. This ensures the flexibility of the local applications.

\begin{algorithm}
	\renewcommand{\algorithmicrequire}{\textbf{Server executes:}}
	\renewcommand{\algorithmicensure}{\textbf{ClientUpdate$(j, \bar{w}, \{u_{l}^{0}\})$:}}
	\caption{VHFL training procedure}
	\label{vhfltp}
	\begin{algorithmic}
		\REQUIRE
		\STATE
		Initialize $w_{0}$, $\bar{w}$
		
		\FOR{each global epoch initialized at $t_{c}=\left \lfloor \frac{t}{E_{l}} \right \rfloor E_{l}$}
		\STATE
		$\mathbb{P}\leftarrow $ (random set of $K$ clients)
		\STATE
		Process $u_{l}^{0}=h_{w_{0}}(x_{l}^{0})$, $l\in I_{j}$, $j\in \mathbb{P}$
		\STATE
		Transmit $\{u_{l}^{0}|l\in I_{j}\}$ and $\bar{w}^{t_{c}}$ to the corresponding client $j$ for all $j \in \mathbb{P}$
		\FOR{each client $j\in \mathbb{P}$ in parallel}
		\STATE
		$w_{j}^{t_{c}+E_{l}}$, $\bigtriangledown_{u^{0}}f_{j}(w_{0}, w_{j})\leftarrow$ ClientUpdate$(j, \bar{w}^{t_{c}}, \{u_{l}^{0}\})$
		\ENDFOR
		\STATE
		$\bar{w}\leftarrow\bar{w}^{t_{c}+E_{l}}=\frac{N}{K}\sum_{j\in \mathbb{P}}q_{j}w_{j}^{t_{c}+E_{l}}$
		\STATE
		$	w_{0}\leftarrow w_{0}^{t_{c}+E_{l}}=w_{0}^{t_{c}}-\eta_{0}^{t_{c}}\sum_{j \in \mathbb{P}}\bigtriangledown_{w_{0}}\bigtriangledown_{u^{0}}f_{j}(w_{0}, w_{j})$
		\ENDFOR
	
		\ENSURE
		\STATE
		$w_{j}\leftarrow\bar{w}$
		\STATE 
		$\mathbb{B}\leftarrow$ (Split $\{x_{l}, y_{l}\}$, $\{u_{l}^{0}\}$ $(l\in I_{j})$ into batches of size $B$)
		\FOR{each local epoch from $t_{c}+1$ to $t_{c}+E_{l}$}
		\FOR{batch $b\in\mathbb{B}$}
		\STATE
		$w_{j}\leftarrow w_{j}-\eta^{t}\bigtriangledown_{w_{j}} f_{j}(w_{0}, w_{j})$
		\STATE
		Record $\{\bigtriangledown_{u_{l}^{0}}f_{j}(w_{0}, w_{j})| l\in I_{j}\}$ for each local epoch
		\ENDFOR
		\ENDFOR
		\STATE
		Average $\{\bigtriangledown_{u_{l}^{0}}f_{j}(w_{0}, w_{j})\}$ among local epochs and aggregate the gradients as a batch $\bigtriangledown_{u^{0}}f_{j}(w_{0}, w_{j})$
		\STATE
		Upload $w_{j}$ and $\bigtriangledown_{u^{0}}f_{j}(w_{0}, w_{j})$ to the server
		
	\end{algorithmic}
\end{algorithm}

\section{Convergence Analysis}
In Section \ref{vhfl_tr}, we have discussed the training procedure of VHFL in detail. In this section, considering the non-i.i.d. data distribution among clients, we will analyze the convergence of VHFL in the convex and non-convex cases. Beforehand, the basic assumptions from the literature and the non-i.i.d. metric for data distribution will be first introduced. Then the convergence bound will be derived and discussed in detail. 
\subsection{Basic assumptions}
\subsubsection{Assumptions on loss function}
The assumptions on L-smooth for smoothness and $\mu$-Polyak-Lojasiewicz (PL) conditions for model convexity are given as follows.
\begin{assumption}\label{L_smooth}
	(L-smooth) The loss function $f(.)$ satisfies
	\begin{align}
		f(\overrightarrow{y}) \leq f(\overrightarrow{x})+(\overrightarrow{y}-\overrightarrow{x})^{T}\bigtriangledown f(\overrightarrow{x}) + \frac{L}{2}\left \| \overrightarrow{y}-\overrightarrow{x} \right \|^{2},
	\end{align}
	where $\overrightarrow{x}$ and $\overrightarrow{y}$ represent a combination of multiple parameters.
\end{assumption}

\begin{assumption}\label{P_L_condition}
	($\mu$-P-L condition) The loss function $f(.)$ in FL training satisfies the following general extension of the $\mu$-strongly convex property
	\begin{align}
		|| \bigtriangledown f(\overrightarrow{x}) ||^{2} \geq 2\mu[f(\overrightarrow{x})-f^{*}],
	\end{align}
	where $f^{*}$ is the optimal loss.
\end{assumption}
\begin{remark}\label{remark_base_assump}
	In analysis of non-convex cases, we will only take Assumption \ref{L_smooth}, while Assumption \ref{P_L_condition} will only be considered for the convex cases. 
	Note that if viewing $f^{*}$ as the local optimum, the assumption could still work for non-convex models. Therefore, the convergence bound for convex cases can still fit for models such as neural networks. 
\end{remark}
\subsubsection{Non-i.i.d. metric}
Due to different local environments, the local observations among clients can be varied in probability distributions. Such a property can largely affect the convergence, which must be considered with the metric in the analytic form.

\begin{definition}\label{define_noniid}
	\cite{yin2018gradient, haddadpour2019convergence}
	Given client set $\mathbb{N}$ with weights $\{q_{j}\}$ and $\{\bigtriangledown f_{j}(w_{0},w_{j})\}$ as their gradients, the metric $\lambda$ for non-i.i.d. extent is defined as
	\begin{equation}\label{bound_non_iid}
		\frac{\sum_{j=1}^{N}q_{j}\left \| \bigtriangledown f_{j}(w_{0}, w_{j}) \right \|^{2}}{\left \| \sum_{j=1}^{N}q_{j}\bigtriangledown f_{j}(w_{0}, w_{j}) \right \|^{2}}=\wedge \leq \lambda.
	\end{equation}
\end{definition}

\begin{remark}
	It is obvious to see that $\lambda \geq 1$. Under such metric, $\lambda=1$ represents the i.i.d. case. The metric $\lambda$ reflects the divergence of the stochastic gradients among the clients.
\end{remark}

\subsection{Convergence bound}

\begin{theorem}\label{conv_bd}
	Denoting the learning rate as $\eta_{t} \propto O(\frac{1}{t})$, the total number of global epochs as $T_{g}=\left \lfloor \frac{t}{E_{l}} \right \rfloor$ and an arbitrary global epoch as $t_{g}$ ($t_{g}\leq T_{g}$), then under Assumption \ref{L_smooth} and Definition \ref{define_noniid}, the VHFL training in the non-convex case converges as
	\begin{align}\label{theorem_non_convex}
		&\frac{1}{T_{g}}\sum_{t_{g}=0}^{T_{g}-1}||\bigtriangledown f(\bar{w}^{t_{g}}, w_{0}^{t_{g}})||^{2}\leq
		2\frac{f(\bar{w}^{0}, w_{0}^{0})-f^{*}}{\sqrt{T_{g}}\sqrt{E_{l}}}\notag \\
		&+\frac{ L\sqrt{E_{l}}}{\sqrt{T_{g}}}[E_{l}\sigma_{0}^{2}+\frac{\sigma^{2}}{K}(\frac{1}{E_{l}}+(\lambda-1)L)+(\lambda-1)LE_{l}G^{2}].
	\end{align}

	Further considering Assumption \ref{P_L_condition}, then the convergence bound of VHFL in the convex case is
	\begin{align}\label{theorem_convex}
		&E[F(w_{0}^{t_{g}}, \bar{w}^{t_{g}})]-f^{*}\leq \frac{1}{T_{g}}\frac{2L}{\mu^{2}}\notag\\
		&[E_{l}\sigma_{0}^{2}+\frac{\sigma^{2}}{K}(\frac{1}{E_{l}}+(\lambda-1)L)+(\lambda-1)LE_{l}G^{2}+\frac{f_{0}G^{2}}{4E_{l}}].
	\end{align}

	Therein, the gradient $||\bigtriangledown f(\bar{w}^{t_{g}}, w_{0}^{t_{g}})||^{2}$ is short for $||\bigtriangledown_{\bar{w}} f(\bar{w}^{t_{g}}, w_{0}^{t_{g}})||^{2}+||\bigtriangledown_{w_{0}} f(\bar{w}^{t_{g}}, w_{0}^{t_{g}})||^{2}$ and $f_{0}$ is proportional to $\bigtriangledown f(\bar{w}^{0}, w_{0}^{0})$. The corner mark represents the gradient at the $t_{g}$-th global epoch and $f(\bar{w}^{0}, w_{0}^{0})$ is the initial loss. The parameter $G^{2}$ is the gradient upper bound and $\sigma^{2}$, $\sigma_{0}^{2}$ are the stochastic gradient variance of $\bigtriangledown_{\bar{w}}f$, $\bigtriangledown_{w_{0}}f$. The proof is shown in Appendix \ref{conv_proof}.
\end{theorem}

In the non-convex case, the bound (\ref{theorem_non_convex}) indicates that the gradients of VHFL converges at the rate $O(\frac{1}{\sqrt{T_{g}}})$. As known, the back proporgation can only ensure that the local optimization problem in non-convex. As the gradients approaches $0$, the result will surely reach a stationary point, which can be viewed as a solution in the non-convex case. In fact, the training of deep neural networks is typically a non-convex case. However, we could ensure the accuracy using methods such as random initialization, enlarging data samples, etc. 

In the convex case, the bound (\ref{theorem_convex}) shows that the loss function converges towards the global optimum at the rate $O(\frac{1}{T_{g}})$. The optimization of the convex problem is rather simple with a single stationary point, which leads to a direct bound with respect to $f^{*}$. In the training of deep neural networks, $f^{*}$ can be viewed as a local optimum. Thus, such bound still maintains its effectiveness to assist the design. 

The hyper-parameters in the background system can largely affect the convergence speed as shown in (\ref{theorem_non_convex}) and (\ref{theorem_convex}). Therein, more collected models in each global epoch results in a larger $K$. Thus, the errors from the non-i.i.d. distribution ($\lambda$) and stochastic variance ($\sigma^{2}$) can be naturally reduced. A larger local epoch leads to more sufficient training on the local data sets, which enhances the overall convergence speed. However, in cases with a high non-i.i.d. distribution ($\lambda$) or large stochastic variance ($\sigma_{0}^{2}$) in the global set, the performance can be worse. Note that the non-i.i.d. distribution could largely affect the convergence. In some extreme cases, $\lambda$ can even approach infinity, which will definitely lead to the failure of convergence. Thus, we should try to reduce the non-i.i.d. distribution in practice. A common operation is to distribute some public data among clients, so that the uploaded models can enjoy a larger similarity in general. 

Compared with HFL, VHFL introduces an additional central stochastic variance ($\sigma_{0}^{2}$). From Theorem \ref{conv_bd}, increasing $E_{l}$ will enlarge the effects of $\sigma_{0}^{2}$. From the view of central training, too much local training may cause large asynchronism and enlarge the variance from the central data sets. In this case, the theorem is fitted with common sense. From an implementation perspective, we can enrich the central set to reduce $\sigma_{0}^{2}$. Then VHFL could guarantee a similar convergence rate as the general HFL training.

\section{Network analysis for VHFL implementation}\label{network_vhfl}
The joint training of the central agency and distributed clients counts on interactions among the global model and local models. However, due to the random flow burden and link conditions, the network is typically unreliable. For services of VHFL, the packet loss and delay in network transmission will degrade the model aggregation. Note that the TCP will retransmit the lost packets, which results in a larger delay. In this case, the packet loss condition can be covered by increasing the network delay. To cope with the unknown transmission time, the central agency will set an upper-bound for delay as $T_{p}$. Any packet arriving latter than $T_{p}$ will be lost. Denoting the average success rate of packet transmission as $\gamma$, then the network will actually collect $K_{\gamma}=K\gamma$ models in each global epoch. Thus, the parameter $K$ in Theorem \ref{conv_bd} should be modified as $K_{\gamma}=K\gamma$. Then (\ref{theorem_non_convex}) for the non-convex cases should be modified as
\begin{align}\label{FLN1}
	&\frac{1}{T_{g}}\sum_{t_{g}=0}^{T_{g}-1}||\bigtriangledown f(\bar{w}^{t_{g}}, w_{0}^{t_{g}})||^{2}\leq
	2\frac{f(\bar{w}^{0}, w_{0}^{0})-f^{*}}{\sqrt{T_{g}}\sqrt{E_{l}}}\notag \\
	&+\frac{ L\sqrt{E_{l}}}{\sqrt{T_{g}}}[E_{l}\sigma_{0}^{2}+\frac{\sigma^{2}}{K\gamma}(\frac{1}{E_{l}}+(\lambda-1)L)+(\lambda-1)LE_{l}G^{2}].
\end{align}
The bound (\ref{theorem_convex}) for the convex cases will be
\begin{align}\label{FLN2}
	&E[f(w_{0}^{t_{g}}, \bar{w}^{t_{g}})]-f^{*}=\frac{1}{T_{g}}\frac{2L}{\mu^{2}}[E_{l}\sigma_{0}^{2}\notag \\
	&+\frac{\sigma^{2}}{K\gamma}(\frac{1}{E_{l}}+(\lambda-1)L)+(\lambda-1)LE_{l}G^{2}+\frac{f_{0}G^{2}}{4E_{l}}].
\end{align}
The convergence of VHFL in practical networks can be described in (\ref{FLN1}) and (\ref{FLN2}), where the model aggregator can suffer from the network loss due to the diminished $K$. Therefore, we should be able to control $\gamma$ so that the aggregated amount $K$ can be guaranteed for VHFL in practical networks. As stated in Section \ref{basic_model}, the network delay is modeled as an $M/G/1$ queuing system. In this section, considering the delay upper-bound $T_{p}$, we will give an analysis of $\gamma$ for VHFL implementation. We will then answer how we should assign the communication module considering the requirements of VHFL training.
  
The network transmission rate is composed of $\mu_{1}$ (idle states) and $\mu_{2}$ (busy states). In practice, the arriving rate $\lambda_{N}$ and $\alpha_{1}$, $\alpha_{2}$, $\mu_{1}$, $\mu_{2}$ in (\ref{delay_prob}) can be measured from the public network with an estimated value. Then an analysis of $\gamma$ can be obtained using the following theorem. 

\begin{figure}[!h]
	\centering
	\includegraphics[width=0.99\columnwidth]{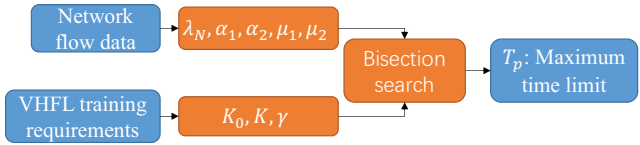}
	\caption{The scheduling procedure for communication in VHFL training.}\label{network_scheduling}
\end{figure}

\begin{theorem}\label{delay_loss}
	Assuming the network transmission between an arbitrary client and the center is an $M/G/1$ process, where $\lambda_{N}$, $\alpha_{1}$, $\alpha_{2}$, $\mu_{1}$, $\mu_{2}$ can be measured from the network. If the delay upper-bound in VHFL for aggregation is set as $T_{p}$, then the success rate $\gamma$ is
	\begin{align}\label{estimate_gamma}
		\gamma=1&+\frac{(1-\rho)(\mu_{1,2} s_{1}+\mu_{1}\mu_{2})}{(s_{1}-s_{2})s_{1}}e^{s_{1}T_{p}}\notag\\
		&-\frac{(1-\rho)(\mu_{1,2} s_{2}+\mu_{1}\mu_{2})}{(s_{1}-s_{2})s_{2}}e^{s_{2}T_{p}},
	\end{align}
	where $\mu_{1,2}=\alpha_{1}\mu_{1}+\alpha_{2}\mu_{2}$ and $\rho=\alpha_{1}\frac{\lambda_{N}}{\mu_{1}}+\alpha_{2}\frac{\lambda_{N}}{\mu_{2}}$. $s_{1}$ and $s_{2}$ are given as
	\begin{align}\label{s1_2}
		&s_{1, 2}=\frac{1}{2}[(\lambda_{N}-\mu_{1}-\mu_{2})\underline{+}\notag\\
		&\sqrt{\mu_{1}^{2}+\mu_{2}^{2}+\lambda_{N}^{2}-2\mu_{1}\mu_{2}+2\lambda_{N}\mu_{1}+2\lambda_{N}\mu_{2}-4\mu_{1,2}\lambda_{N}}].
	\end{align}
	The proof is provided in Appendix \ref{network_proof}.
\end{theorem}

Theorem \ref{delay_loss} provides a method to estimate the failure rate in model aggregations due to network conditions and delay setting. From Theorem \ref{conv_bd}, the training convergence improves with $K$ in an inversely proportional manner. Then if $K$ is not large enough, the training can be seriously affected. In practice, we can obtain an appropriate $K$ from experience so that VHFL can gain reasonable convergence. Then by facing such unreliable network, how can we ensure that we can actually achieve $K$? 

The scheduling of VHFL communication links through the public internet is summarized in Fig. \ref{network_scheduling}. 
Assume that the network permits $K_{0}$ clients to upload models due to the scheduled capability for VHFL communication, we should ensure $\gamma\geq \frac{K}{K_{0}}$ to guarantee the training convergence. In public networks, the transmission rate can be estimated by existing flow data by experience. By setting $\gamma=\frac{K}{K_{0}}$, we could solve (\ref{estimate_gamma}) by bisection search to get $T_{p}$. For model aggregation in VHFL, we could let the central trainer wait for time $T_{p}$ to collect distributed feedback before training. Here, $T_{p}$ is a guide from theoretical model. The practical system may perform slight adjustment around $T_{p}$ through experiments.

\section{Experiments}
\subsection{Basic settings}
We test the VHFL on the flow data in a practical network among a range of communities. The target is to predict the network throughput of each cell. In this case, the cells are the distributed clients with observations from the local flow records. Meanwhile, the central flow data reflecting the features of several neighboring cells are stored by the central agency as the global feature. Each data sample is a vector composed of features from the flow records. The data features are composed of the typical network parameter settings involving the average user trafic, average and maximum number of active users, small packet ratio, etc. Assisted by the central counterparts, the features among distributed cells form the distributed data sets for FL. The samples in all cells and the central counterparts are divided into a training set and test set. The training loss and the inference test accuracy are recorded in each training mode.

Due to properties of the data features, we employ a multilayer perceptron model (MLP) in the experiments. In distributed cells, the local features are first processed by an MLP model. Then the output feature is concatenated with the globally processed feature and processed by a subsequent MLP model to predict the throughput. At the central agency, a similar MLP model is applied to process the global observations. The results are shared with the local counterparts and concatenated in the local model. 

\begin{figure}[!h]
	\centering
	\includegraphics[width=0.9\columnwidth]{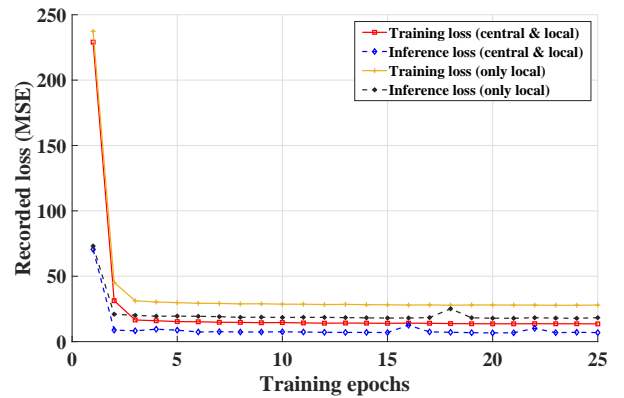}
	\caption{The effectiveness of introducing the central features in central training.}\label{central_local_compare}
\end{figure}

\subsection{Cloud training benchmark}
We shall employ the cloud training on the local and global features of all cells as a basic benchmark. Therein, the cloud training concatenates the local and central models as a whole, following the same structure as VHFL. Meanwhile, we also perform the cloud training by excluding the central feature to validate its effectiveness. The comparison results are shown in Fig. \ref{central_local_compare}, where the cloud training with the combined features obviously outperforming the results based on the local features. This means that the central features are important in this task, which is a typical scenario for VHFL.

\subsection{VHFL performance}
\begin{figure*}[tbp]
\centering
\subfigure[Training loss comparison.]{
\label{val_vhfl_1}
\includegraphics[width=0.32\textwidth]{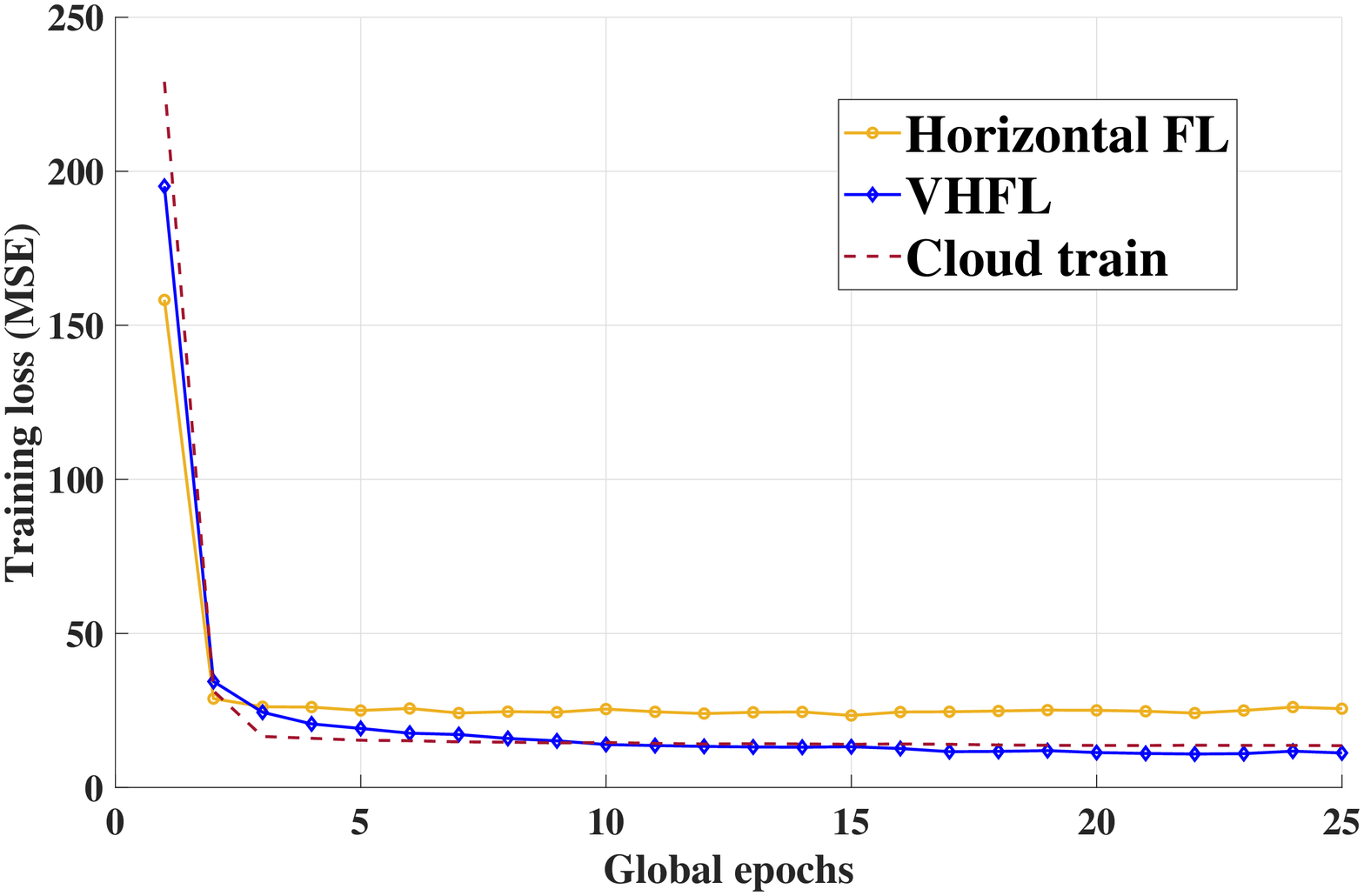}}
\subfigure[Inference test loss comparison.]{
	\label{val_vhfl_2}
	\includegraphics[width=0.32\textwidth]{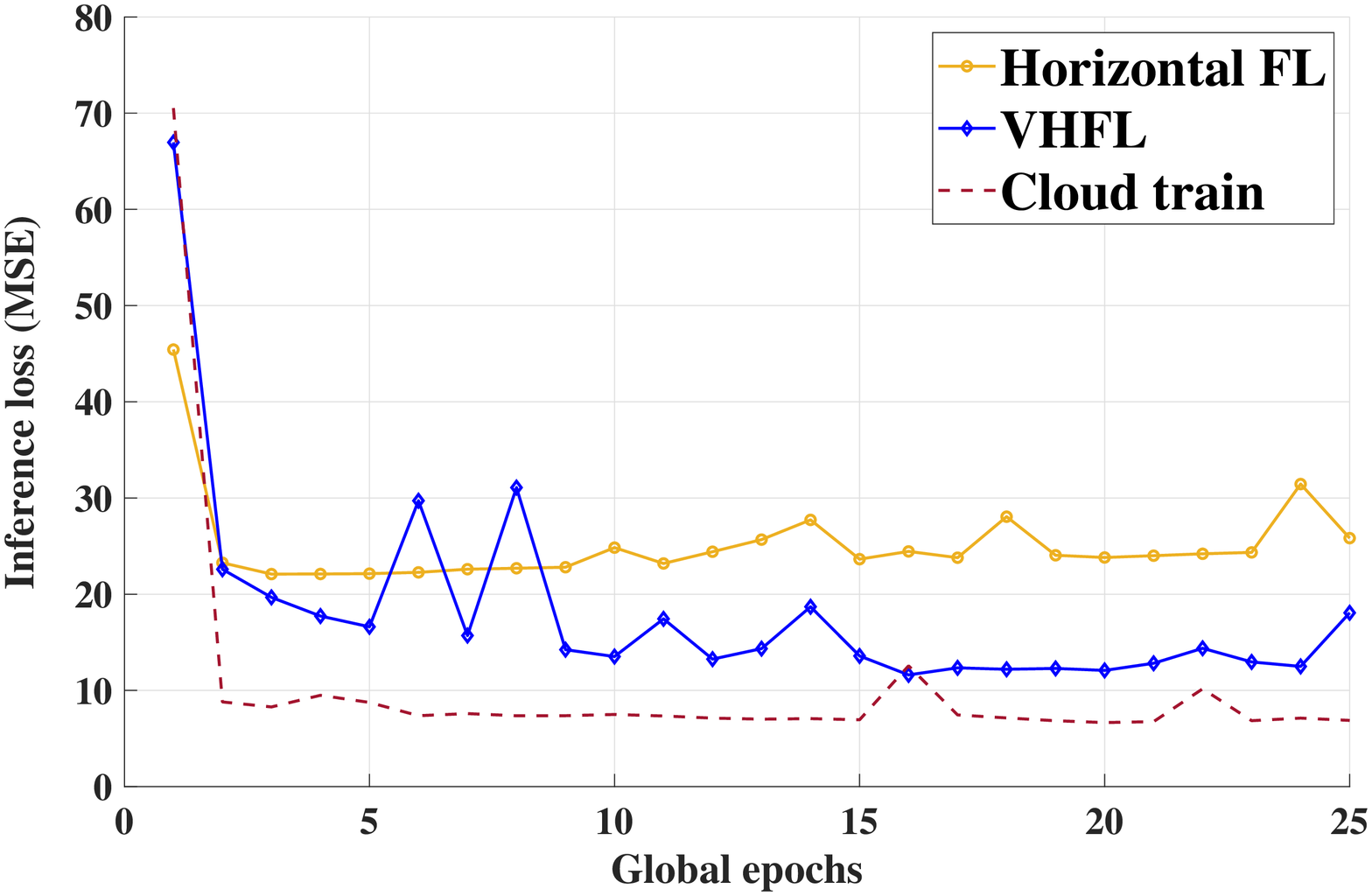}}
\subfigure[Prediction accuracy comparison.]{
	\label{val_vhfl_3}
	\includegraphics[width=0.32\textwidth]{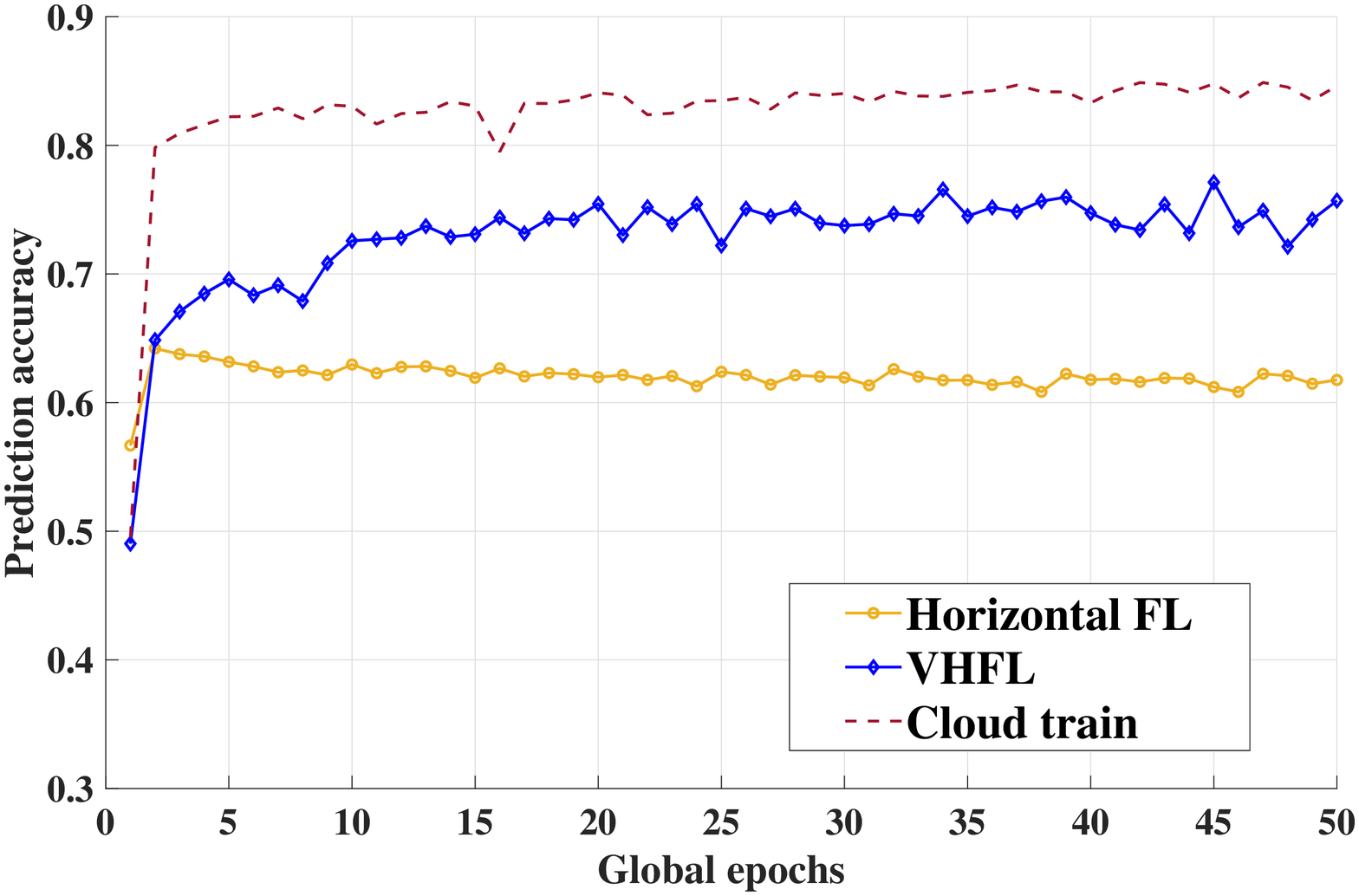}}
\caption{A comparison of central training with all features, VHFL and Horizontal FL in experiments (Training loss, Inference test loss, and prediction accuracy).}
\end{figure*}

Here, we apply VHFL among the cells and compare the prediction loss with the Horizontal FL and cloud training benchmark. In the HFL mode, the training is only based on the local features where the center only performs model aggregation for distributed cells. Fig. \ref{val_vhfl_1} shows the training convergence where VHFL could achieve a similar speed as HFL. Meanwhile, with the central feature assistance, the achieved training loss in VHFL is apparently improved. Fig. \ref{val_vhfl_2} shows the MSE test loss for validation set and Fig. \ref{val_vhfl_3} shows the corresponding prediction error ratio. As shown in the curves, the model trained by VHFL performs well from a generalization ability. Thus, employing the central feature in FL can significantly improve the model accuracy for applications in the network.

\begin{figure}[!h]
	\centering
	\subfigure[Convergence trend versus $K$.]{
		\label{val_K}
		\includegraphics[width=0.4\textwidth]{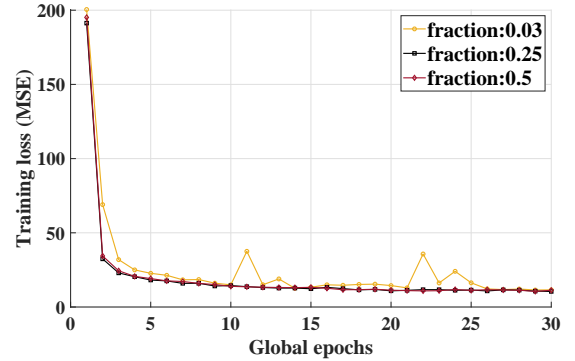}}
	\subfigure[Convergence trend versus $E_{l}$.]{
		\label{val_E}
		\includegraphics[width=0.4\textwidth]{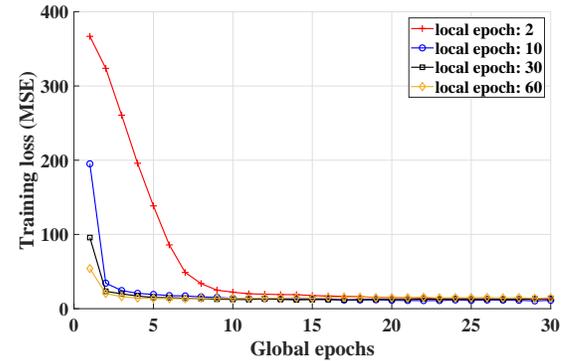}}
	\caption{The convergence tendency of VHFL training with $K$ and $E_{l}$.}
\end{figure}

\subsection{VHFL convergence}
As observed in Theorem \ref{conv_bd}, $K$ and $E_{l}$ are important parameters for training. Here, we observe the convergence tendency of VHFL with these parameters, as shown in Fig. \ref{val_K} and Fig. \ref{val_E}. In Fig. \ref{val_K}, the fraction represents $\frac{K}{N}$. While $K$ is small, increasing $K$ could enhance the speed and stability in training. If $K$ gets large enough, the effects get smaller. Here we improved the non-i.i.d. data sets by adding a small public set to the distributed cells. If the non-i.i.d. property gets worse, the effects of $K$ should be larger. In Fig. \ref{val_E}, the convergence speed increases with the local epoch. In this sense, we could increase $E_{l}$ in VHFL to reduce the necessary communication rounds. However, the communication gain will get smaller as $E_{l}$ keeps increasing. In practice, setting $E_{l}=20$ may be a good choice.

\subsection{Delay setting for VHFL implementation}
\begin{figure}[!h]
	\centering
	\subfigure[Tendency of $\gamma$ versus $T_{p}$.]{
		\label{gamma_tp}
		\includegraphics[width=0.4\textwidth]{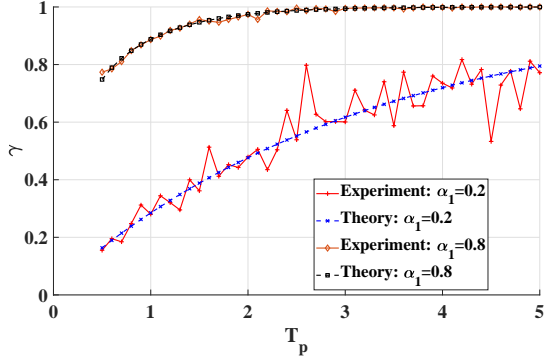}}
	\subfigure[Required $T_{p}$ versus $\gamma$.]{
		\label{require_tp_gamma}
		\includegraphics[width=0.4\textwidth]{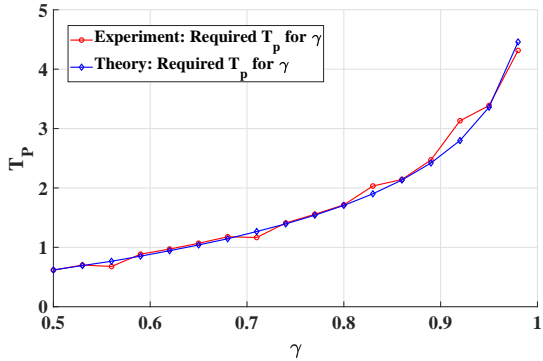}}
	\caption{The success uploading rate with the delay upper-bound in VHFL.}
\end{figure}

In Section \ref{network_vhfl}, we discussed the delay setting considering the requirement of model collection in VHFL. In practice, since each uploading client and even the message is lost, will consume communication resources, the uploading amount may be restricted. Meanwhile, VHFL will require a certain minimum value of $K$ by experience. Thus, the designer needs to control the delay upper-bound $T_{p}$ so that $K$ uploaded messages will be received in general. By setting $\lambda_{N}=2$, $\mu_{1}=8$, and $\mu_{2}=2$, we simulate the random network queuing delay as a benchmark. As shown in Fig. \ref{gamma_tp}, the theoretical analysis fits well with the experimental results. As $\alpha_{1}$ gets smaller, the network will become more stressed. Then we may need a much larger $T_{p}$ to achieve the same $\gamma$. Meanwhile, we set $\alpha_{1}=\alpha_{2}=0.5$ and change the target $\gamma$ to see the required $T_{p}$. As shown in Fig. \ref{require_tp_gamma}, the required $T_{p}$ increases exponentionally with the target $\gamma$. In practice, we could estimate the network conditions and jointly design the system parameters under theoretical guidance.

\section{Conclusion}
In future 6G communication, the intelligence assisted by distributed learning will be a major engine to assist the IoT applications. From the networking perpestive, how to share the experience while protecting the data security should be an important consideration. The proposal of FL addressed the problem of sharing experience among distributed clients. However, how to combine the central information in distributed training still remains to be solved. For security concern, the global data in the network center can not be directly broadcast. However, if we can utilize such information without direct data transmission, the obtained intelligence will be significantly enhanced. 

In this paper, we discussed how to introduce the use of global observations in FL. By combining the training procedure of vertical and horizontal FL, we developed a collaborative training method called Vertical-Horizontal Federated learning (VHFL) where a central agency can train a model to process the global observations. Through theoretical analysis, we developed an analytic bound for the VHFL training loss and showed that the central model will converge with the federal model and achieve a similar convergence rate as conventional HFL. Considering the practical implementation of VHFL, we discussed the effects of network delay and proposed the theoretical principle to schedule the waiting time in model collection. Finally, extensive experimental results were used to validated the effectiveness and convergence of VHFL and were shown to fit well with the theoretical results. In practical network, VHFL can be applied to applications where the global information is effective for distributed intelligent clients.

%\bibliographystyle{ieeetr}
%\bibliography{bibfile}

\appendices
\section{Convergence analysis}\label{conv_proof}
In the following, we will denote $\bigtriangledown_{w_{0}} f(\bar{w}^{t}, w_{0}^{t})$ as $g_{0}^{t}$, $\bigtriangledown_{w_{j}} f(w_{j}^{t}, w_{0}^{t})$ as $g_{j}^{t}$ and write $\bigtriangledown_{\bar{w}} f(\bar{w}^{t}, w_{0}^{t})$ as $\bigtriangledown_{w}f^{t}$ for short. Meanwhile, $\widetilde{g}_{0}^{t}$ and $\widetilde{g}_{j}^{t}$ represent the corresponding stochastic gradients for an arbitrary training slot. Since central training only occurs at $t=t_{c}=\left \lfloor \frac{t}{E_{l}} \right \rfloor E_{l}$, the expectation of $\widetilde{g}_{0}^{t}$ is in the following $E[\widetilde{g}_{0}^{t}]=\widehat{g}_{0}^{t}$, where $\widehat{g}_{0}^{t}=g_{0}^{t}$ for $t=t_{c}$ and $\widehat{g}_{0}^{t}=0$, $\widetilde{g}_{0}^{t}=0$ for $t\neq t_{c}$. $\widehat{g}_{0}^{t}$ here is a notation considering the absense of central training in local epochs. The variance of these stochastic gradients are assumed to be $E[||\widehat{g}_{0}^{t}-\widetilde{g}_{0}^{t}||^{2}]\leq \sigma_{0}^{2}$, $E[||g_{j}^{t}-\widetilde{g}_{j}^{t}||^{2}]\leq \sigma^{2}$ and the gradient bound is supposed to be $E[||g_{j}^{t}||^{2}]\leq G^{2}$.

The learning rate of the central and local training is $\eta_{0}^{t}$, $\eta^{t}$. In model aggregation, we suppose an arbitrary client $j$ is chosen with probability $q_{j}$. In this case, (\ref{wbar_vhfl}) can be equvally written as $\bar{w}^{t}=\frac{1}{K}\sum_{j\in \mathbb{P}}w_{j}^{t}$. Note that the $\bar{w}^{t}$ here is a virtually aggregated weight. The distributed $w_{j}^{t}$ is set as $\bar{w}^{t}$ only at $t=t_{c}=\left \lfloor \frac{t}{E_{l}} \right \rfloor E_{l}$. Then from the L-smooth property in Assumption \ref{L_smooth}, the loss function in VHFL training satisfies
\begin{align}\label{basic_conv}
	&E[f(w_{0}^{t+1}, \Bar{w}^{t+1})]\leq E[f(w_{0}^{t}, \Bar{w}^{t})]+\notag\\&E[-\eta_{0}^{t}<g_{0}^{t}, \widetilde{g}_{0}^{t}>-\eta^{t}<\bigtriangledown_{w} f^{t}, \frac{1}{K}\sum_{j\in \mathbb{P}}\widetilde{g}_{j}^{t}> +\notag\\& \frac{L}{2}(\eta_{0}^{t})^{2}||\widetilde{g}_{0}^{t}||^{2} + \frac{L}{2}(\eta^{t})^{2}||\frac{1}{K}\sum_{j\in \mathbb{P}}\widetilde{g}_{j}^{t}||^{2}].
\end{align}

Since, $E[\widetilde{g}_{0}^{t}]=\widehat{g}_{0}^{t}$, then the term with respect to global training is bounded by
\begin{align}\label{bound_0}
	&-\eta_{0}^{t}<g_{0}^{t}, \widehat{g}_{0}^{t}>+\frac{L}{2}(\eta_{0}^{t})^{2}||\widetilde{g}_{0}^{t}||^{2} \leq -\frac{\eta_{0}^{t}}{2}(||g_{0}^{t}||^{2}+||\widehat{g}_{0}^{t}||^{2})\notag\\&+\frac{\eta_{0}^{t}}{2}||g_{0}^{t}-\widehat{g}_{0}^{t}||^{2}+\frac{L}{2}(\eta_{0}^{t})^{2}||\widehat{g}_{0}^{t}||^{2}+\frac{L}{2}(\eta_{0}^{t})^{2}||\widehat{g}_{0}^{t}-\widetilde{g}_{0}^{t}||^{2}.
\end{align}
Bound (\ref{bound_0}) can be obtained using $-ab=\frac{(a-b)^{2}-a^{2}-b^{2}}{2}$ and $E[X^{2}]=(E[X])^{2}+E[(X-E[X])^{2}]$. Then assuming that $E[||\widehat{g}_{0}^{t}-\widetilde{g}_{0}^{t}||^{2}]\leq \sigma_{0}^{2}$ and $\eta_{0}^{t}\leq\frac{1}{L}$, then $\frac{L}{2}(\eta_{0}^{t})^{2}||\widehat{g}_{0}^{t}||^{2}\leq \frac{\eta_{0}^{t}}{2}||\widehat{g}_{0}^{t}||^{2}$. (\ref{bound_0}) is then bounded by
\begin{align}
	&E[-\eta_{0}^{t}<g_{0}^{t}, \widehat{g}_{0}^{t}>+\frac{L}{2}(\eta_{0}^{t})^{2}||\widetilde{g}_{0}^{t}||^{2}] \leq -\frac{\eta_{0}^{t}}{2}||g_{0}^{t}||^{2}\notag\\&+\frac{\eta_{0}^{t}}{2}||g_{0}^{t}-\widehat{g}_{0}^{t}||^{2}+\frac{L}{2}(\eta_{0}^{t})^{2}\sigma_{0}^{2}.
\end{align}
If $t=t_{c}$, $g_{0}^{t}=\widehat{g}_{0}^{t}$, then $-\frac{\eta_{0}^{t}}{2}||g_{0}^{t}||^{2}+\frac{\eta_{0}^{t}}{2}||g_{0}^{t}-\widehat{g}_{0}^{t}||^{2} = -\frac{\eta_{0}^{t}}{2}||g_{0}^{t}||^{2}$ . If $t\neq t_{c}$, then $-\frac{\eta_{0}^{t}}{2}||g_{0}^{t}||^{2}+\frac{\eta_{0}^{t}}{2}||g_{0}^{t}-\widehat{g}_{0}^{t}||^{2} = 0$. Since $t$ evolves to be $t_{c}$ at the period of $E_{l}$, then this becomes equal to $-\frac{\eta_{0}^{t}}{2E_{l}}||g_{0}^{t}||^{2}$ in the long run. Denoting $\frac{\eta_{0}^{t}}{E_{l}}$ as $\widehat{\eta}_{0}^{t}$, then
\begin{align}\label{glo1}
	&-\eta_{0}^{t}<g_{0}^{t}, \widehat{g}_{0}^{t}>+\frac{L}{2}(\eta_{0}^{t})^{2}||\widetilde{g}_{0}^{t}||^{2} \leq -\frac{\widehat{\eta}_{0}^{t}}{2}||g_{0}^{t}||^{2} \notag \\
	&+\frac{LE_{l}^{2}}{2}(\widehat{\eta}_{0}^{t})^{2}\sigma_{0}^{2}.
\end{align}

For the FL part,  $\frac{L}{2}(\eta^{t})^{2}||\frac{1}{K}\sum_{j\in \mathbb{P}}\widetilde{g}_{j}^{t}||^{2}$ in (\ref{basic_conv}) is bounded by
\begin{align}\label{square1}
	&E[\frac{L}{2}(\eta^{t})^{2}||\frac{1}{K}\sum_{j\in \mathbb{P}}\widetilde{g}_{j}^{t}||^{2}] \overset{\textcircled{1}}{=}
	\frac{L}{2}(\eta^{t})^{2}E||\frac{1}{K}\sum_{j\in \mathbb{P}}(g_{j}^{t}-\widetilde{g}_{j}^{t})||^{2}\notag\\
	&+\frac{L}{2}(\eta^{t})^{2}||E[\frac{1}{K}\sum_{j\in \mathbb{P}}\widetilde{g}_{j}^{t}]||^{2}
	\overset{\textcircled{2}}{\leq} \frac{L}{2}(\eta^{t})^{2}\frac{1}{K^{2}}\sum_{j\in \mathbb{P}}E||g_{j}^{t}-\widetilde{g}_{j}^{t}||^{2}\notag\\
	&+\frac{L}{2}(\eta^{t})^{2}||\sum_{j=1}^{N}q_{j}g_{j}^{t}||^{2}
	\overset{\textcircled{3}}{\leq} \frac{L}{2}(\eta^{t})^{2}\frac{\sigma^{2}}{K}+\frac{L}{2}(\eta^{t})^{2}||\sum_{j=1}^{N}q_{j}g_{j}^{t}||^{2},
\end{align}
where $\textcircled{1}$ is true because $E[X^{2}]=(E[X])^{2}+E[(X-E[X])^{2}]$. Since the random gradient error is independent among clients, $E[(g_{j}^{t}-\widetilde{g}_{j}^{t})(g_{i}^{t}-\widetilde{g}_{i}^{t})]=0(i\neq j)$. Meanwhile, due to $E_{\mathbb{P}}[\frac{1}{K}\sum_{j\in \mathbb{P}}g_{j}^{t}]=\sum_{j=1}^{N}q_{j}g_{j}^{t}$ and $E||g_{j}^{t}-\widetilde{g}_{j}^{t}||^{2} \leq \sigma^{2}$, $\textcircled{2}$ and $\textcircled{3}$ hold.
As for $E[-\eta^{t}<\bigtriangledown_{w} f^{t}, \frac{1}{K}\sum_{j\in \mathbb{P}}\widetilde{g}_{j}^{t}>]$ in (\ref{basic_conv}), we have
\begin{align}\label{bound_mul1}
	&-\eta^{t}<\bigtriangledown_{w} f^{t}, E[\frac{1}{K}\sum_{j\in \mathbb{P}}\widetilde{g}_{j}^{t}]>=\notag\\&-\eta^{t}<\bigtriangledown_{w} f^{t}, \frac{1}{K}\sum_{j\in \mathbb{P}}g_{j}^{t}>=-\frac{\eta^{t}}{2}||\bigtriangledown_{w} f^{t}||^{2}\notag\\
	&-\frac{\eta^{t}}{2}||\sum_{j=1}^{N}q_{j}g_{j}^{t}||^{2}+\frac{\eta^{t}}{2}||\bigtriangledown_{w} f^{t}-\sum_{j=1}^{N}q_{j}g_{j}^{t}||^{2}.
\end{align}
Therein, from the L-smooth property as stated in Assumption \ref{L_smooth}, $||\bigtriangledown_{w} f^{t}-\sum_{j=1}^{N}q_{j}g_{j}^{t}||^{2}=||\sum_{j=1}^{N}q_{j}(\bigtriangledown_{w} f_{j}(\bar{w}^{t}, w_{0}^{t})-g_{j}^{t})||^{2}\leq L^{2}\sum_{j=1}^{N}q_{j}||\bar{w}^{t}-w_{j}^{t}||^{2}$. From the basic training process, we have,
\begin{equation}\label{upw1}
	\bar{w}^{t_{c}}=\frac{1}{K}\sum_{j\in \mathbb{P}}w^{t_{c}}_{j},
\end{equation}
\begin{equation}\label{upw2}
	w_{j}^{t}=\bar{w}^{t_{c}}-\sum_{k=t_{c}+1}^{t-1}\eta^{k}\widetilde{g}_{j}^{k},
\end{equation}
\begin{equation}\label{upw3}
	\bar{w}^{t}=\bar{w}^{t_{c}}-\frac{1}{K}\sum_{j\in \mathbb{P}}\sum_{k=t_{c}+1}^{t-1}\eta^{k}\widetilde{g}_{j}^{k}.
\end{equation}

Since $E[\frac{1}{K}\sum_{j\in \mathbb{P}}||\bar{w}^{t}-w_{j}^{t}||^{2}]=\sum_{j=1}^{N}q_{j}||\bar{w}^{t}-w_{j}^{t}||^{2}$, the assessment of $||\bigtriangledown_{w} f^{t}-\sum_{j=1}^{N}q_{j}g_{j}^{t}||^{2}$ is equivalent to the analysis of $\frac{1}{K}\sum_{j\in \mathbb{P}}||\bar{w}^{t}-w_{j}^{t}||^{2}$. From (\ref{upw2}) and (\ref{upw3}), we have
\begin{align}\label{difw1}
	&\frac{1}{K}\sum_{j\in \mathbb{P}}||\bar{w}^{t}-w_{j}^{t}||^{2}
	=\frac{1}{K}\sum_{j\in \mathbb{P}}||\sum_{k=t_{c}+1}^{t-1}\eta^{k}\widetilde{g}_{j}^{k}-\notag \\
	&\frac{1}{K}\sum_{l\in \mathbb{P}}\sum_{k=t_{c}+1}^{t-1}\eta^{k}\widetilde{g}_{l}^{k}||^{2}
	\overset{\textcircled{1}}{=}\sum_{j\in \mathbb{P}}\frac{1}{K}||\sum_{k=t_{c}+1}^{t-1}\eta^{t}\widetilde{g}_{j}^{k}||^{2}-\notag \\
	&||\sum_{j\in \mathbb{P}}\frac{1}{K}\sum_{k=t_{c}+1}^{t-1}\eta^{t}\widetilde{g}_{j}^{k}||^{2}\overset{\textcircled{2}}{\leq}(\lambda-1)||\sum_{j\in \mathbb{P}}\frac{1}{K}\sum_{k=t_{c}+1}^{t-1}\eta^{t}\widetilde{g}_{j}^{k}||^{2}.
\end{align}

Since $E[\widetilde{g}_{j}^{k}]=g_{j}^{k}$, $E[\sum_{j\in \mathbb{P}}\frac{1}{K}\sum_{k=t_{c}+1}^{t-1}\eta^{t}\widetilde{g}_{j}^{k}]=\sum_{j\in \mathbb{P}}\frac{1}{K}\sum_{k=t_{c}+1}^{t-1}\eta^{t}g_{j}^{k}$. Then, because $E[X^{2}]=E[(X-E[X])^{2}]+(E[X])^{2}$, we get
\begin{align}\label{difw2}
	&(\lambda-1)E[||\sum_{j\in \mathbb{P}}\frac{1}{K}\sum_{k=t_{c}+1}^{t-1}\eta^{t}\widetilde{g}_{j}^{k}||^{2}]=(\lambda-1)\notag\\
	&E[||\sum_{j\in \mathbb{P}}\frac{1}{K}\sum_{k=t_{c}+1}^{t-1}\eta^{k}(\widetilde{g}_{j}^{k}-g_{j}^{k})||^{2}+||\sum_{j\in \mathbb{P}}\frac{1}{K}\sum_{k=t_{c}+1}^{t-1}\eta^{k}g_{j}^{k}||^{2}]\notag \\
	&\overset{\textcircled{1}}{=}(\lambda-1)E[\frac{1}{K^{2}}\sum_{j\in \mathbb{P}}\sum_{k=t_{c}+1}^{t-1}(\eta^{k})^{2}||\widetilde{g}_{j}^{k}-g_{j}^{k}||^{2}+\notag \\ 
	&||\sum_{j\in \mathbb{P}}\frac{1}{K}\sum_{k=t_{c}+1}^{t-1}\eta^{k}g_{j}^{k}||^{2}]
	\overset{\textcircled{2}}{\leq}(\lambda-1)E[\frac{1}{K^{2}}\sum_{j\in \mathbb{P}}\sum_{k=t_{c}+1}^{t-1}(\eta^{k})^{2}\notag \\
	&||\widetilde{g}_{j}^{k}-g_{j}^{k}||^{2}+\sum_{j\in \mathbb{P}}\frac{t-t_{c}}{K}\sum_{k=t_{c}+1}^{t-1}(\eta^{k})^{2}||g_{j}^{k}||^{2}].
\end{align}
Since $t-t_{c}<E_{l}$, $E[||\widetilde{g}_{j}^{t}-g_{j}^{t}||^{2}]\leq \sigma^{2}$ and $E[||g_{j}^{t}||^{2}]\leq G^{2}$, (\ref{difw2}) can be bounded by
\begin{align}\label{difw3}
	&(\lambda-1)E[||\sum_{j\in \mathbb{P}}\frac{1}{K}\sum_{k=t_{c}+1}^{t-1}\eta^{t}\widetilde{g}_{j}^{k}||^{2}] \leq (\lambda-1)E[\sum_{j\in \mathbb{P}}\frac{1}{K^{2}}\notag\\
	&\sum_{k=t_{c}+1}^{t_{c}+E_{l}}(\eta^{k})^{2}
	||\widetilde{g}_{j}^{k}-g_{j}^{k}||^{2}+\sum_{j\in \mathbb{P}}\frac{E_{l}}{K}\sum_{k=t_{c}+1}^{t_{c}+E_{l}}(\eta^{k})^{2}||g_{j}^{k}||^{2}]\notag \\
	&\leq (\lambda-1)[\sum_{k=t_{c}+1}^{t_{c}+E_{l}}(\eta^{k})^{2}\frac{\sigma^{2}}{K}+\sum_{k=t_{c}+1}^{t_{c}+E_{l}}(\eta^{k})^{2}E_{l}G^{2}].
\end{align}
Suppose that $(\eta^{k})^{2}$ in (\ref{difw3}) satisfies $(\eta^{k})^{2}\leq \eta^{t}$, (\ref{difw1}), (\ref{difw2}) and (\ref{difw3}) finally leads to
\begin{align}\label{difw4}
	&E[\frac{1}{K}\sum_{j\in \mathbb{P}}||\bar{w}^{t}-w_{j}^{t}||^{2}]\leq (\lambda-1)E_{l}\eta^{t}[\frac{\sigma^{2}}{K}+E_{l}G^{2}].
\end{align}
Also, (\ref{bound_mul1}) and (\ref{difw4}) leads to
\begin{align}\label{difw5}
	&-\eta^{t}<\bigtriangledown_{w} f^{t}, E[\frac{1}{K}\sum_{j\in \mathbb{P}}\widetilde{g}_{j}^{t}]>\leq -\frac{\eta^{t}}{2}||\bigtriangledown_{w} f^{t}||^{2}\notag \\
	&-\frac{\eta^{t}}{2}||\sum_{j=1}^{N}q_{j}g_{j}^{t}||^{2}+(\lambda-1)E_{l}\frac{(\eta^{t})^{2}}{2}L^{2}[\frac{\sigma^{2}}{K}+E_{l}G^{2}].
\end{align}
By combining (\ref{difw5}) and (\ref{square1}), we get
\begin{align}\label{difw6}
	&E[-\eta^{t}<\bigtriangledown_{w} f^{t}, \frac{1}{K}\sum_{j\in \mathbb{P}}\widetilde{g}_{j}^{t}>+\frac{L}{2}(\eta^{t})^{2}||\frac{1}{K}\sum_{j\in \mathbb{P}}\widetilde{g}_{j}^{t}||^{2}]\notag \\
	&\leq -\frac{\eta^{t}}{2}||\bigtriangledown_{w} f^{t}||^{2}+\frac{L}{2}(\eta^{t})^{2}\frac{\sigma^{2}}{K}+(\lambda-1)E_{l}\frac{(\eta^{t})^{2}}{2}L^{2}\notag\\
	&[\frac{\sigma^{2}}{K}+E_{l}G^{2}]-\frac{\eta^{t}}{2}||\sum_{j=1}^{N}q_{j}g_{j}^{t}||^{2}+\frac{L}{2}(\eta^{t})^{2}||\sum_{j=1}^{N}q_{j}g_{j}^{t}||^{2}.
\end{align}
Suppose that $\eta^{t}\leq \frac{1}{L}$, then (\ref{difw6}) leads to
\begin{align}\label{difw6}
	&E[-\eta^{t}<\bigtriangledown_{w} f^{t}, \frac{1}{K}\sum_{j\in \mathbb{P}}\widetilde{g}_{j}^{t}>+\frac{L}{2}(\eta^{t})^{2}||\frac{1}{K}\sum_{j\in \mathbb{P}}\widetilde{g}_{j}^{t}||^{2}]\notag \\
	&\leq -\frac{\eta^{t}}{2}||\bigtriangledown_{w} f^{t}||^{2}+\frac{L}{2}(\eta^{t})^{2}\frac{\sigma^{2}}{K}\notag\\
	&+(\lambda-1)E_{l}\frac{(\eta^{t})^{2}}{2}L^{2}[\frac{\sigma^{2}}{K}+E_{l}G^{2}].
\end{align}
Then from (\ref{basic_conv}), (\ref{glo1}) and (\ref{difw6}) and set $\widehat{\eta}_{0}^{t}=\eta^{t}$, the convergence bound gets to
\begin{align}\label{conv_f}
	&E[f(w_{0}^{t+1}, \Bar{w}^{t+1})]\leq E[f(w_{0}^{t}, \Bar{w}^{t})]-\frac{\eta^{t}}{2}||\bigtriangledown_{w} f^{t}||^{2}\notag \\
	&-\frac{\widehat{\eta}_{0}^{t}}{2}||g_{0}^{t}||^{2}+\frac{LE_{l}^{2}}{2}(\widehat{\eta}_{0}^{t})^{2}\sigma_{0}^{2}+\frac{L}{2}(\eta^{t})^{2}\frac{\sigma^{2}}{K}\notag \\
	&+(\lambda-1)E_{l}\frac{(\eta^{t})^{2}}{2}L^{2}[\frac{\sigma^{2}}{K}+E_{l}G^{2y}]\notag \\
	&=E[f(w_{0}^{t}, \Bar{w}^{t})]-\frac{\eta^{t}}{2}(||\bigtriangledown_{w} f^{t}||^{2}
	+||\bigtriangledown_{w_{0}} f^{t}||^{2})\notag \\&+\frac{LE_{l}^{2}}{2}(\eta^{t})^{2}\sigma_{0}^{2}+\frac{L}{2}(\eta^{t})^{2}\frac{\sigma^{2}}{K}\notag \\
	&+(\lambda-1)E_{l}\frac{(\eta^{t})^{2}}{2}L^{2}[\frac{\sigma^{2}}{K}+E_{l}G^{2}]\notag \\
	&=E[f(w_{0}^{t}, \Bar{w}^{t})]-\frac{\eta^{t}}{2}||\bigtriangledown f(\bar{w}^{t}, w_{0}^{t})||^{2}+\frac{LE_{l}^{2}}{2}(\eta^{t})^{2}\sigma_{0}^{2}\notag \\&
	+\frac{L}{2}(\eta^{t})^{2}\frac{\sigma^{2}}{K}+(\lambda-1)E_{l}\frac{(\eta^{t})^{2}}{2}L^{2}[\frac{\sigma^{2}}{K}+E_{l}G^{2}].
\end{align}

\subsection{Non-convex case}
The bound in (\ref{conv_f}) can be transformed as
\begin{align}\label{nocvx1}
	&E[f(w_{0}^{t+1}, \bar{w}^{t+1})]\leq E[f(w_{0}^{t}, \bar{w}^{t})]-\frac{\eta^{t}}{2}||\bigtriangledown f(\bar{w}^{t}, w_{0}^{t})||^{2}+\notag \\
	&\frac{LE_{l}}{2}(\eta^{t})^{2}[E_{l}\sigma_{0}^{2}+\frac{\sigma^{2}}{K}(\frac{1}{E_{l}}+(\lambda-1)L)+(\lambda-1)LE_{l}G^{2}].
\end{align}
Transforming both sides of (\ref{nocvx1}) by $\frac{1}{T}\sum_{t=0}^{T-1}[.]$, it gets to
\begin{align}\label{nocvx2}
	&\frac{1}{T}\sum_{t=0}^{T-1}[E[f(\bar{w}^{t+1}, w_{0}^{t+1})]-E[f(\bar{w}^{t}, w_{0}^{t})]]\leq\notag \\ &-\frac{1}{T}\sum_{t=0}^{T-1}\frac{\eta^{t}}{2}||\bigtriangledown f(\bar{w}^{t}, w_{0}^{t})||^{2}+\frac{LE_{l}}{2}[E_{l}\sigma_{0}^{2}+\frac{\sigma^{2}}{K}(\frac{1}{E_{l}}\notag \\
	&+(\lambda-1)L)+(\lambda-1)LE_{l}G^{2}]\frac{1}{T}\sum_{t=0}^{T-1}(\eta^{t})^{2}.
\end{align}
Then denoting the optimum as $f^{*}$, we have
\begin{align}\label{nocvx3}
	&\frac{1}{T}\sum_{t=0}^{T-1}\frac{\eta^{t}}{2}||\bigtriangledown f(\bar{w}^{t}, w_{0}^{t})||^{2}\leq
	\frac{f(\bar{w}^{0}, w_{0}^{0})-f^{*}}{T}+\frac{LE_{l}}{2}[E_{l}\sigma_{0}^{2}\notag \\
	&+\frac{\sigma^{2}}{K}(\frac{1}{E_{l}}+(\lambda-1)L)+(\lambda-1)LE_{l}G^{2}]\frac{1}{T}\sum_{t=0}^{T-1}(\eta^{t})^{2}.
\end{align}
Taking $\eta^{t}=\eta$, (\ref{nocvx3}) leads to
\begin{align}\label{nocvx4}
	&\frac{1}{T}\sum_{t=0}^{T-1}||\bigtriangledown f(\bar{w}^{t}, w_{0}^{t})||^{2}\leq
	2\frac{f(\bar{w}^{0}, w_{0}^{0})-f^{*}}{\eta T}+\eta LE_{l}[E_{l}\sigma_{0}^{2}\notag \\
	&+\frac{\sigma^{2}}{K}(\frac{1}{E_{l}}+(\lambda-1)L)+(\lambda-1)LE_{l}G^{2}].
\end{align}
Then taking $\eta=O(\frac{1}{\sqrt{T}})$, we finally have
\begin{align}\label{nocvx5}
	&\frac{1}{T}\sum_{t=0}^{T-1}||\bigtriangledown f(\bar{w}^{t}, w_{0}^{t})||^{2}\leq
	2\frac{f(\bar{w}^{0}, w_{0}^{0})-f^{*}}{\sqrt{T}}+\frac{ LE_{l}}{\sqrt{T}}[E_{l}\sigma_{0}^{2}\notag \\
	&+\frac{\sigma^{2}}{K}(\frac{1}{E_{l}}+(\lambda-1)L)+(\lambda-1)LE_{l}G^{2}].
\end{align}
Denoting the total global epoch as $T_{g}=\frac{T}{E_{l}}$ and the current round of the global epoch as $t_{g}=\frac{t_{c}}{E_{l}}$, then it can be directly observed from (\ref{nocvx5}) that
\begin{align}\label{nocvx6}
	&\frac{1}{T_{g}}\sum_{t_{g}=0}^{T_{g}-1}||\bigtriangledown f(\bar{w}^{t_{g}}, w_{0}^{t_{g}})||^{2}\leq
	2\frac{f(\bar{w}^{0}, w_{0}^{0})-f^{*}}{\sqrt{T_{g}}\sqrt{E_{l}}}\notag \\
	&+\frac{ L\sqrt{E_{l}}}{\sqrt{T_{g}}}[E_{l}\sigma_{0}^{2}+\frac{\sigma^{2}}{K}(\frac{1}{E_{l}}+(\lambda-1)L)+(\lambda-1)LE_{l}G^{2}].
\end{align}
\subsection{Convex case}
For the convex training model, Assumption \ref{P_L_condition} holds. From (\ref{conv_f}), we have
\begin{align}\label{fcvx1}
	&E[f(w_{0}^{t+1}, \bar{w}^{t+1})]\leq E[f(w_{0}^{t}, \bar{w}^{t})]-\frac{\eta^{t}}{2}||\bigtriangledown f(\bar{w}^{t}, w_{0}^{t})||^{2}+\notag \\
	&\frac{LE_{l}}{2}(\eta^{t})^{2}[E_{l}\sigma_{0}^{2}+\frac{\sigma^{2}}{K}(\frac{1}{E_{l}}+(\lambda-1)L)+(\lambda-1)LE_{l}G^{2}].
\end{align}
Then, from Assumption \ref{P_L_condition}, (\ref{fcvx1}) leads to
\begin{align}\label{cvx1}
	&E[f(w_{0}^{t+1}, \bar{w}^{t+1})]\leq E[f(w_{0}^{t}, \bar{w}^{t})]-\mu \eta^{t}(E[f(w_{0}^{t}, \bar{w}^{t})]-f^{*})\notag \\
	&+\frac{LE_{l}}{2}(\eta^{t})^{2}[E_{l}\sigma_{0}^{2}+\frac{\sigma^{2}}{K}(\frac{1}{E_{l}}+(\lambda-1)L)+(\lambda-1)LE_{l}G^{2}].
\end{align}
By subtracting $f^{*}$ on both sides of (\ref{cvx1}), we obtain
\begin{align}\label{cvx2}
	&E[f(w_{0}^{t+1}, \bar{w}^{t+1})]-f^{*}\leq (1-\mu \eta^{t})(E[f(w_{0}^{t}, \bar{w}^{t})]-f^{*})\notag \\
	&+\frac{LE_{l}}{2}(\eta^{t})^{2}[E_{l}\sigma_{0}^{2}+\frac{\sigma^{2}}{K}(\frac{1}{E_{l}}+(\lambda-1)L)+(\lambda-1)LE_{l}G^{2}].
\end{align}
(\ref{cvx2}) can then be written as
\begin{align}\label{cvx3}
	E[f(w_{0}^{t+1},& \bar{w}^{t+1})]-f^{*}\notag \\
	&\leq (1-\mu \eta^{t})(E[f(w_{0}^{t}, \bar{w}^{t})]-f^{*})+(\eta^{t})^{2}M,
\end{align}
where $M$ denotes $\frac{LE_{l}}{2}[E_{l}\sigma_{0}^{2}+\frac{\sigma^{2}}{K}(\frac{1}{E_{l}}+(\lambda-1)L)+(\lambda-1)LE_{l}G^{2}]$. By observing the bound in (\ref{cvx3}), the training loss will converge with a diminished learning rate. By taking $\eta^{t}=\frac{\beta}{t+l_{0}} (\beta>0, l_{0}>0)$, we will prove that the training loss converges in the form $E[f(w_{0}^{t}, \bar{w}^{t})]\leq \frac{A}{t+l_{0}}$. Using the induction method, we suppose that the convergence bound holds for $t=1$. For simplicity, $E[f(w_{0}^{t}, \bar{w}^{t})]$ is denoted as $\Delta_{t}$ for short. Thus, assuming $\Delta_{t} \leq \frac{A}{t+l_{0}}$ holds, (\ref{cvx3}) leads to
\begin{align}\label{cvx4}
	\Delta_{t+1}&\leq (1-\mu \eta^{t})\Delta_{t}+(\eta^{t})^{2}M \notag \\
	&\leq (1-\frac{\beta \mu}{t+l_{0}})\frac{A}{t+l_{0}}+\frac{\beta^{2}M}{(t+l_{0})^{2}}\notag \\
	&= \frac{t+l_{0}-\beta\mu}{(t+l_{0})^{2}}A+\frac{\beta^{2}M}{(t+l_{0})^{2}}\notag \\
	&=\frac{t+l_{0}-1}{(t+l_{0})^{2}}A+\frac{\beta^{2}M-(\beta\mu-1)A}{(t+l_{0})^{2}}.
\end{align}
Then assuming that $\beta\mu>1$ and $A\geq \frac{\beta^{2}M}{\beta\mu-1}$, then (\ref{cvx4}) leads to
\begin{align}\label{cvx5}
	\Delta_{t+1}\leq \frac{t+l_{0}-1}{(t+l_{0})^{2}}A\leq \frac{A}{t+l_{0}+1}.
\end{align}
By considering the initial condition, $\Delta_{1}\leq \frac{A}{1+l_{0}}$. Thus, the parameter $A$ should be ${\rm max}\{\frac{\beta^{2}M}{\beta\mu-1}, \Delta_{1}(l_{0}+1)\}$ with $\Delta_{t}\leq \frac{A}{t+l_{0}}$. Specifically, it is reasonable to choose $\beta=\frac{2}{\mu}$. Then the result could be derived as
\begin{align}\label{cvx6}
	E[f(w_{0}^{t},& \bar{w}^{t})]-f^{*}\leq \frac{1}{t+l_{0}}(\frac{4}{\mu^{2}}M+(l_{0}+1)\Delta_{1}).
\end{align}
By the properties of L-smooth, $\bigtriangleup_{1} \leq \frac{L}{2}|| w_{0}-w^{*} ||^{2}$. Given the $\mu$-strongly convexity, $\mu||w_{0}-w^{*}|| \leq ||\bigtriangledown f(w_{0}, w_{0})-\bigtriangledown f^{*}||$. Since $\bigtriangledown f^{*}=0$, we get $||w_{0}-w^{*}||^{2}\leq \frac{1}{\mu^{2}}\left \| \bigtriangledown f(w_{0}, w_{0}) \right \|^{2}$. Suppose that $\left \| \bigtriangledown f(w_{0}) \right \|^{2} \leq f_{0}G^{2}$, then $\left \| w_{0}-w^{*} \right \|^{2} \leq \frac{f_{0}}{\mu^{2}}G^{2}$. Therefore, $\bigtriangleup_{1} \leq \frac{Lf_{0} G^{2}}{2\mu^{2}}$. By setting $l_{0}=0$ and taking in $T_{g}=\frac{T}{E_{l}}$, then if we train $T_{g}$ global epochs, the convergence bound is
\begin{align}\label{cvx7}
	&E[f(w_{0}^{t_{g}}, \bar{w}^{t_{g}})]-f^{*}\leq \frac{1}{E_{l}T_{g}}[\frac{2LE_{l}}{\mu^{2}}[E_{l}\sigma_{0}^{2}+\frac{\sigma^{2}}{K}(\frac{1}{E_{l}}+\notag \\
	&(\lambda-1)L)+(\lambda-1)LE_{l}G^{2}]+\frac{Lf_{0}G^{2}}{2\mu^{2}}]=\frac{1}{T_{g}}\frac{2L}{\mu^{2}}\notag\\
	&[E_{l}\sigma_{0}^{2}+\frac{\sigma^{2}}{K}(\frac{1}{E_{l}}+(\lambda-1)L)+(\lambda-1)LE_{l}G^{2}+\frac{f_{0}G^{2}}{4E_{l}}].
\end{align}

\section{Network analysis}\label{network_proof}
For an $M/G/1$ system, $N_{t_{n}^{+}}$ denotes the queuing length when packet $n$ departures and $V_{n}$ denotes the number of arriving packets while packet $n$ is being handled. Thus, it is natural that $N_{t_{n}^{+}}$ updates as $N_{t_{n+1}^{+}}=N_{t_{n}^{+}}+V_{n+1}-1$ for $N_{t_{n}^{+}}>0$ and $N_{t_{n}^{+}}=V_{n+1}$ for $N_{t_{n}^{+}}=0$. Here, the target is to analyze $Q=\lim_{n \to \infty } N_{t_{n}^{+}}$ to further derive the distribution of the network delay. We have
\begin{equation}\label{NA1}
	E[N_{t_{n+1}^{+}}]=E[N_{t_{n}^{+}}]-E[\Delta_{N_{t_{n}^{+}}}]+E[V_{n+1}],
\end{equation}
where $\Delta_{N_{t_{n}^{+}}}=1$ for $N_{t_{n}^{+}}>0$ and $\Delta_{N_{t_{n}^{+}}}=0$ for $N_{t_{n}^{+}}=0$. The probability generating function of $N_{t_{n}^{+}}$ is defined as
\begin{align}\label{NA2}
	Q_{n}(z)=\sum_{k=0}^{\infty}P[N_{t_{n}^{+}}=k]z^{k}=E[z^{N_{t_{n}^{+}}}].
\end{align}
Considering the steady state of the network, the probability generating function can be defined as
\begin{align}\label{NA3}
	Q(z)=\lim_{n \to \infty }Q_{n}(z)=\sum_{k=0}^{\infty}P[Q=k]z^{k}=E[z^{Q}].
\end{align}
Since $V_{n+1}$ and $N_{t_{n}^{+}}$ are independent from each other, (\ref{NA1}) leads to
\begin{align}\label{NA4}
	Q_{n+1}(z)=E[z^{N_{t_{n+1}^{+}}}]=E[z^{N_{t_{n}^{+}}-\Delta_{N_{t_{n}^{+}}}}]E[z^{V_{n+1}}].
\end{align}
Since $V_{n}$ is not related with the index $n$, $E[z^{V_{n+1}}]$ can be given as follows:
\begin{align}\label{NA5}
	V(z)&=E[z^{V}]=\sum_{k=0}^{\infty}P[v=k]z^{k}\notag\\
	&=\sum_{k=0}^{\infty}[\int_{0}^{\infty }\frac{(\lambda_{N} x)^{k}}{k!}e^{-\lambda_{N} x}b(x) dx]z^{k}\notag\\
	&=\int_{0}^{\infty }e^{-\lambda_{N} x}\sum_{k=0}^{\infty}\frac{(\lambda_{N} xz)^{k}}{k!} b(x) dx \notag\\
	&=\int_{0}^{\infty }e^{-\lambda_{N} x+\lambda_{N} xz}b(x) dx=\int_{0}^{\infty }e^{-\lambda_{N}(1-z)x}b(x) dx.
\end{align}
Denoting the Laplace transmission of $b(x)$ as $B(s)$, then $V(z)=B(\lambda_{N}-\lambda_{N} z)$. $E[z^{N_{t_{n}^{+}}-\Delta_{N_{t_{n}^{+}}}}]$ in (\ref{NA4}) can then be derived as follows: 
\begin{align}\label{NA6}
	&E[z^{N_{t_{n}^{+}}-\Delta_{N_{t_{n}^{+}}}}]	= \sum_{k=0}^{\infty}P[N_{t_{n}^{+}}=k]z^{k-\Delta_{k}}\notag \\
	&=P[N_{t_{n}^{+}}=0]+\sum_{k=1}^{\infty}P[N_{t_{n}^{+}}=k]z^{k-1}\notag \\
	&=P[N_{t_{n}^{+}}=0]+\frac{1}{z}\sum_{k=0}^{\infty}P[N_{t_{n}^{+}}=k]z^{k}-\frac{1}{z}P[N_{t_{n}^{+}}=0]\notag \\
	&=P[N_{t_{n}^{+}}=0]+\frac{N_{t_{n}^{+}}(z)-P[N_{t_{n}^{+}}=0]}{z}.
\end{align}
By substituting (\ref{NA3}) into (\ref{NA6}), we get
\begin{align}\label{NA7}
	Q(z)=V(z)[P[Q=0]+\frac{Q(z)-P[Q=0]}{z}].
\end{align}
It is a basic result in queuing theory that $P[Q=0]=1-\rho$, where $\rho=\alpha_{1}\frac{\lambda_{N}}{\mu_{1}}+\alpha_{2}\frac{\lambda_{N}}{\mu_{2}}$. Combined with (\ref{NA5}), the queuing length distribution can be obtained as
\begin{align}\label{NA8}
	Q(z)=B(\lambda_{N}-\lambda_{N} z)\frac{(1-\rho)(1-z)}{B(\lambda_{N}-\lambda_{N} z)-z}.
\end{align}
From (\ref{NA5}), there is $V(z)=B(\lambda_{N}-\lambda_{N} z)$. Here, $B(z)$ denotes the serving time and $V(z)$ represents the number of arriving packets while a packet is being handled. Let $D(z)$ be the network delay, then the number of arriving packets during this period should be the queuing length. Thus, $Q(z)=D(\lambda_{N}-\lambda_{N} z)$ also holds. Together with (\ref{NA8}) and setting $s=\lambda_{N}-\lambda_{N} z$, then the Laplace transform of the network delay is
\begin{align}\label{NA9}
	D(s)=B(s)\frac{(1-\rho)s}{s-\lambda_{N}+\lambda_{N} B(s)}.
\end{align}
Since the probability distribution of $B(t)$ is $B(t)=\alpha_{1}\mu_{1}e^{-\mu_{1}t}+\alpha_{2}\mu_{2}e^{-\mu_{2}t}$, its Laplace transform can be obtained as
\begin{align}\label{NA10}
	B(s)=\frac{\alpha_{1}\mu_{1}}{s+\mu_{1}}+\frac{\alpha_{2}\mu_{2}}{s+\mu_{2}}.
\end{align}
By substituting (\ref{NA10}) into (\ref{NA9}) and defining the notation $\mu_{1,2}=\alpha_{1}\mu_{1}+\alpha_{2}\mu_{2}$, we obtain
\begin{align}\label{NA11}
	&D(s)=\notag\\
	&\frac{(1-\rho)(\mu_{1,2} s+\mu_{1}\mu_{2})}{s^{2}+(\mu_{1}+\mu_{2}-\lambda_{N})s+(\mu_{1,2}\lambda_{N}+\mu_{1}\mu_{2}-\lambda_{N}\mu_{1}-\lambda_{N}\mu_{2})}\notag\\
	&=\frac{(1-\rho)(\mu_{1,2} s+\mu_{1}\mu_{2})}{(s-s_{1})(s-s_{2})},
\end{align}
where $s_{1}$ and $s_{2}$ are the zero points of the denominator in (\ref{NA11}) given as follows:
\begin{align}\label{NA12}
	&s_{1, 2}=\frac{1}{2}[(\lambda_{N}-\mu_{1}-\mu_{2})\underline{+}\notag\\
	&\sqrt{\mu_{1}^{2}+\mu_{2}^{2}+\lambda_{N}^{2}-2\mu_{1}\mu_{2}+2\lambda_{N}\mu_{1}+2\lambda_{N}\mu_{2}-4\mu_{1,2}\lambda_{N}}].
\end{align}
From (\ref{NA11}), the probability distribution of the network delay can be derived from the inverse Laplace transformation as
\begin{align}\label{NA13}
	W(t)=(1-\rho)[\frac{(\mu s_{1}+\mu_{1}\mu_{2})}{s_{1}-s_{2}}e^{s_{1}t}-\frac{(\mu s_{2}+\mu_{1}\mu_{2})}{s_{1}-s_{2}}e^{s_{2}t}],
\end{align}
where $t\geq 0$.
Thus, assuming that $T_{p}$ is the admitted delay upper-bound, then the success rate of the packet transmission is
\begin{align}\label{NA14}
	&\gamma=1-\int_{T_{p}}^{\infty}W(t) dt=1+\frac{(1-\rho)(\mu_{1,2} s_{1}+\mu_{1}\mu_{2})}{(s_{1}-s_{2})s_{1}}e^{s_{1}T_{p}}\notag\\
	&-\frac{(1-\rho)(\mu_{1,2} s_{2}+\mu_{1}\mu_{2})}{(s_{1}-s_{2})s_{2}}e^{s_{2}T_{p}}.
\end{align}
This completes the analysis of $\gamma$ as a function of $T_{p}$.

\vfill

\end{document}